\documentclass[11pt]{article}
\usepackage{acl}
\usepackage{times}
\usepackage{latexsym}
\usepackage[T1]{fontenc}
\usepackage[utf8]{inputenc}
\usepackage{microtype}
\usepackage{inconsolata}
\usepackage{paralist}
\usepackage{CJKutf8}
\usepackage{booktabs}
\usepackage{multirow}
\usepackage{amsmath}
\usepackage{amssymb}
\usepackage[table]{xcolor}
\usepackage{graphicx}
\usepackage{placeins}
\usepackage{float}
\usepackage{hyperref}
\usepackage[bottom,flushmargin]{footmisc}
\newcommand\blfootnote[1]{%
  \begingroup
  \renewcommand\thefootnote{}\footnote{#1}%
  \addtocounter{footnote}{-1}%
  \endgroup
}
\hypersetup{colorlinks=true, linkcolor=blue, citecolor=blue, urlcolor=blue}

\definecolor{drop}{RGB}{155,110,180}   
\definecolor{mag}{RGB}{110,110,120}

\newcommand{\cis}[2]{{\tiny[#1,\,#2]}}

\newcommand{\rA}[1]{\cellcolor{mag!30}$#1$}
\newcommand{\rB}[1]{\cellcolor{mag!14}$#1$}
\newcommand{\hA}[1]{\cellcolor{drop!45}$#1$}
\newcommand{\hB}[1]{\cellcolor{drop!22}$#1$}

\newcommand{\dA}[1]{\cellcolor{drop!45}#1}   % >= 0.15
\newcommand{\dB}[1]{\cellcolor{drop!22}#1}   % 0.05--0.15

\newcommand{\dH}[1]{\cellcolor{drop!55}#1}   % >= 0.30
\newcommand{\dM}[1]{\cellcolor{drop!35}#1}   % 0.15--0.30
\newcommand{\dL}[1]{\cellcolor{drop!18}#1}   % 0.05--0.15

\newcommand{\mA}[1]{\cellcolor{mag!30}#1}
\newcommand{\mB}[1]{\cellcolor{mag!14}#1}

\title{Lexical Perturbations Disrupt LLM Reasoning:
An Empirical Study of Attention Diversion}

\author{
  \textbf{Jiaqian Zhu}$^{1}$ \quad
  \textbf{Yang Zhang}$^{2,*}$ \quad
  \textbf{Junhua Ding}$^{2}$ \quad
  \textbf{Xiaowei Yu}$^{1,*}$ \\
  $^{1}$Missouri University of Science and Technology, Rolla, MO, USA \\
  $^{2}$University of North Texas, Denton, TX, USA \\
  \texttt{\{jzbmn, xych8\}@mst.edu} \\
  \texttt{\{Yang.Zhang, Junhua.Ding\}@unt.edu} \\
  $^{*}$Corresponding authors
}

\begin{document}

\maketitle

\blfootnote{Accepted to the 2026 Conference on Empirical Methods in Natural
Language Processing (EMNLP 2026), Main Conference.}

\begin{abstract}
Large Language Models (LLMs) achieve strong reasoning performance, but their
robustness to realistic lexical corruption remains poorly understood. We
evaluate four open-weight instruction-tuned models and frontier models across
four reasoning benchmarks under keyboard noise, character swaps, and filler
insertion. Character-level perturbations substantially degrade accuracy,
especially on multi-step reasoning tasks, while filler insertion has little
effect. We trace this asymmetry to \textit{Attention Diversion}: lexical corruption
fragments subword tokenization, and the resulting fragments attract
disproportionate attention mass, concentrated in middle and final transformer
layers. Length-matched controls confirm that fragmentation, not prompt length,
drives the loss. A factorial intervention then shows why the damage is hard to
undo: fragmentation corrupts token content and attention allocation together,
and the two are coupled. Restoring clean attention while the content remains
corrupted is actively harmful, restoring content alone is insufficient, and
only restoring both recovers a substantial share of the gap. This coupling
explains why inference-time strategies, including chain-of-thought prompting,
spell-checking, self-repair, and stronger repair models, fail to consistently
recover performance: each addresses one channel at a time. Code and data are
available at
\url{https://github.com/Jiaqian-Janelle/Attention-Diversion}.
\end{abstract}

% -------------------------------Section 1：Introduction ------------------------------------%

\section{Introduction}
\label{sec:intro}

Large Language Models (LLMs) have achieved substantial gains on reasoning
benchmarks
\cite{brown2020gpt3,openai2023gpt4,wei2022chain,llama3,qwen35,mistral},
yet remain brittle under small surface-form changes. Minor lexical
corruptions that preserve meaning, such as mistyped keys or
adjacent-character swaps, can markedly degrade accuracy despite being
transparent to human readers. In contrast, filler insertions such as
\textit{``um''} or \textit{``you know''} leave performance nearly
unchanged, even though they increase prompt length. This asymmetry
suggests the failure is not a consequence of longer prompts or semantic
distortion, but of how surface-form corruption is represented.

Prior work has studied character-level noise in neural NLP systems
\cite{belinkov2018synthetic,pruthi2019combating,li2019textbugger,ebrahimi2018hotflip},
but three gaps remain. First, it concentrates on classification and
translation, leaving reasoning-intensive tasks underexplored. Second, it
alters surface form and prompt length together, leaving the two
confounded even though only the former changes subword token identity.
Third, it documents accuracy losses without connecting them to changes
in internal processing.

We address these gaps through a controlled empirical study of lexical
perturbations in LLM reasoning. We evaluate four open-weight
instruction-tuned models across BoolQ, PIQA, HellaSwag, and GSM8K,
and further test larger-scale, harder, non-English, and frontier-model
settings. Across these settings, character-level perturbations
consistently degrade performance, with the largest losses on multi-step
mathematical reasoning, whereas filler insertion has little effect.

\begin{figure}[t]
\centering
\includegraphics[width=\columnwidth]{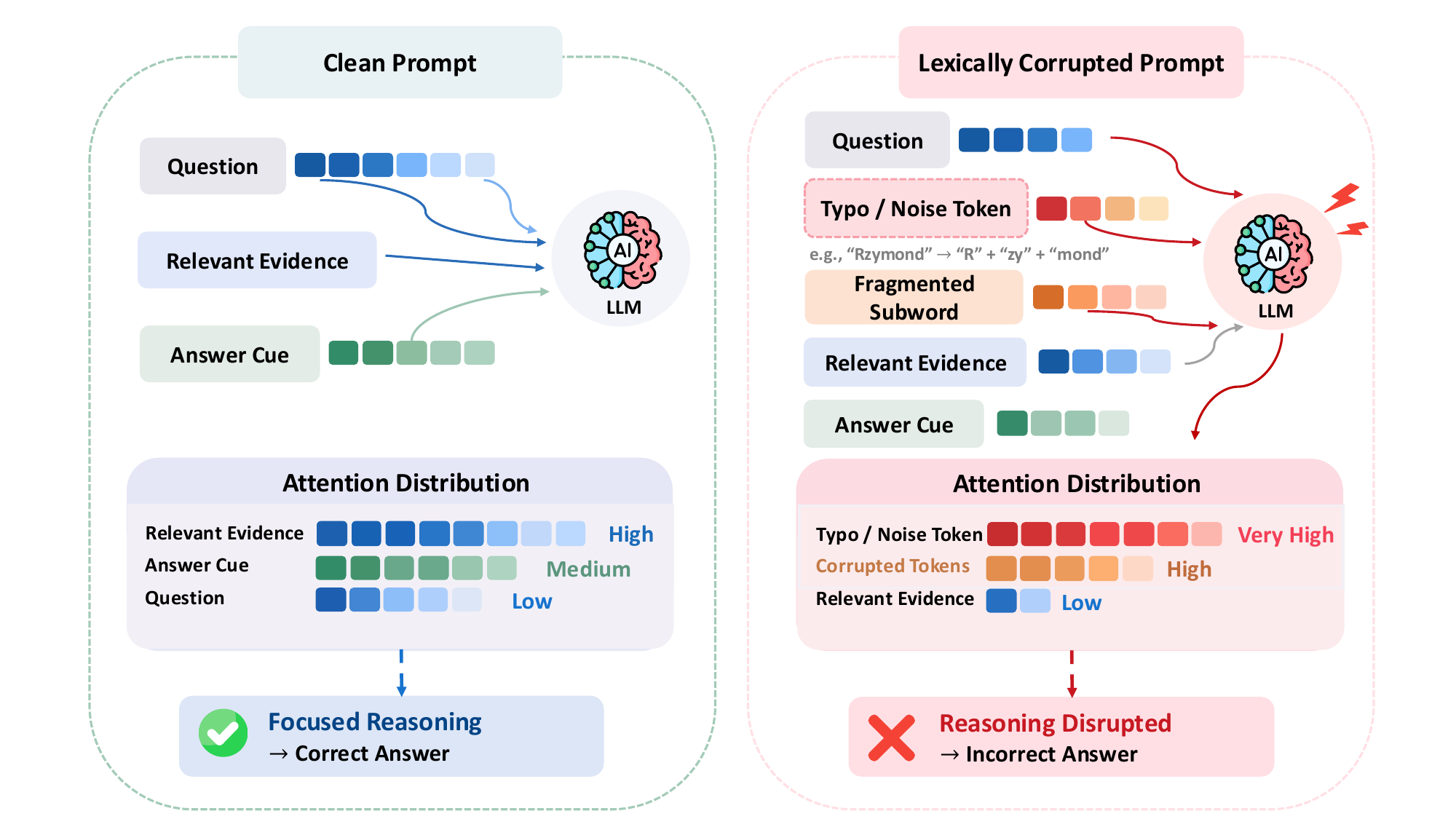}
\caption{Illustration of Attention Diversion under lexical corruption.
In clean prompts, attention concentrates on relevant evidence and answer
cues. After perturbation, corrupted token fragments absorb a
disproportionate share of attention mass.}
\label{fig:attention_hijack_concept}
\end{figure}

To explain this asymmetry, we identify \textbf{Attention Diversion}.
Character-level corruption fragments subword tokenization, producing
rare or unusual token pieces that attract disproportionate attention
mass during inference. As illustrated in
Figure~\ref{fig:attention_hijack_concept}, corrupted fragments redirect
attention away from task-relevant evidence and answer cues. Layer-wise
analysis further shows that this diversion concentrates in middle and
final transformer layers, where contextual representations are composed
and mapped toward output predictions.

Our central claim is that lexical fragility in reasoning is driven
primarily by tokenizer fragmentation rather than prompt length alone.
Length-matched controls, per-example regression, and representation-level
interventions support this account. 
A factorial intervention isolates what fragmentation damages: it disrupts the
identity of the tokens the model reads and the allocation of attention over
them at the same time, and these two effects cannot be separated. This
interdependence also accounts for our mitigation results: 
common inference-time strategies, including chain-of-thought
prompting, spell-checking, and LLM self-repair, do not consistently recover
performance, because each acts on one side of the problem alone.

Our contributions are:
\begin{compactitem}
\item We systematically evaluate lexical robustness across four reasoning
benchmarks, four open-weight model families, frontier models, and realistic
perturbation types.

\item We identify \textbf{Attention Diversion} and, through a factorial
intervention, show that the diverted attention is not independently
manipulable: it is functionally bound to the corrupted content it addresses,
so neither channel can be repaired on its own.

\item We trace harm to the irreplaceability of displaced evidence rather
than to diversion magnitude, with numeric tokens dominating GSM8K failures.
We then bound repair, finding that a tokenizer-level defense detects
fragmentation reliably yet cannot invert it.
\end{compactitem}

% -------------------------------Section 2：Lexical Perturbations in LLM Reasoning ------------------------------------%
\section{Lexical Perturbations in LLM Reasoning}
\label{sec:perturbations}

We use \textit{lexical perturbation} to denote surface-form changes
that preserve the intended meaning of a prompt while altering how the
input is written. Such perturbations arise naturally from typing errors,
speech-to-text transcription, and noisy prompt construction. We focus on
three realistic perturbation types, illustrated in
Table~\ref{tab:perturb_examples}.

\paragraph{Perturbation types.}
\textbf{Keyboard noise} replaces a character with an adjacent QWERTY key
(e.g., \textit{reason} $\rightarrow$ \textit{reaspn}), simulating fast or
imprecise typing. \textbf{Character swap} (typoswap) transposes adjacent
characters within a word (e.g., \textit{problem} $\rightarrow$
\textit{porblem}), modeling common spelling errors. \textbf{Filler
insertion} adds conversational disfluencies such as \textit{``um''} or
\textit{``you know''} at natural pause points, reflecting spoken or
voice-transcribed input.

\paragraph{Perturbation severity.}
We apply perturbations at four rates
$r \in \{0.05, 0.10, 0.20, 0.30\}$, where $r$ denotes the fraction of
words perturbed. For keyboard noise and character swap, one randomly
chosen character within each selected word is modified. Filler insertion
adds disfluency phrases without corrupting existing lexical items. This
design lets us compare character-level lexical corruption against prompt
length inflation.

\begin{table}[t]
\centering
\small
\caption{Illustrative examples of the three perturbation types applied to a
single base prompt, with the perturbed span in \textcolor{red}{red}.}
\label{tab:perturb_examples}
\setlength{\tabcolsep}{4pt}
\renewcommand{\arraystretch}{1.2}
\begin{tabular}{@{}l p{0.63\columnwidth}@{}}
\toprule
\textbf{Perturbation} & \textbf{Prompt} \\
\midrule
Clean & \textit{Which option best explains why the ice melted faster on the
metal surface?} \\
Keyboard noise & \textit{Which option best exp\textcolor{red}{k}ains why the
ice melted faster on the metal surface?} \\
Typoswap & \textit{Which option best explains why the ice
melt\textcolor{red}{de} faster on the metal surface?} \\
Filler & \textit{\textcolor{red}{Um,} which option best explains why the ice
melted faster on the metal surface?} \\
\bottomrule
\end{tabular}
\end{table}

\paragraph{Expected representation effect.}
Character-level perturbations can break learned subword merges, causing
a word to be split into rare or unusual token fragments. In contrast,
filler insertion mainly adds familiar tokens while preserving the
tokenization of the original prompt. This contrast motivates our
mechanistic analysis in Section~\ref{sec:mechanism}, where we test
whether tokenizer fragmentation induces Attention Diversion and
reasoning failure.

% -------------------------------Section 3：Experimental Setup ------------------------------------%
\section{Experimental Setup}
\label{sec:experimental_setup}

\subsection{Datasets}

We evaluate lexical robustness on four reasoning benchmarks:
BoolQ \cite{boolq} for passage-based yes/no question answering,
PIQA \cite{piqa} for physical commonsense reasoning,
HellaSwag \cite{hellaswag} for context continuation, and
GSM8K \cite{gsm8k} for multi-step mathematical reasoning. Together,
they span reading comprehension, commonsense reasoning, contextual
inference, and math reasoning.

\subsection{Models}

Our main experiments cover four open-weight instruction-tuned models:
Llama-3.1-8B-Instruct \cite{llama3},
Mistral-7B-Instruct-v0.3 \cite{mistralv03},
Qwen3.5-9B \cite{qwen35},
and Gemma-2-9B-IT \cite{gemma2}.
These models have comparable scale (7B--9B) but differ in model family,
tokenizer design, and training pipeline, allowing us to test whether
lexical fragility generalizes across architectures. We additionally
evaluate GPT-4o \cite{gpt4o} and GPT-5.4 \cite{gpt54}, the latter under both \texttt{none} and \texttt{high} \texttt{reasoning\_effort} settings, for prediction-level
robustness.
Frontier-model results
are reported in Appendix~\ref{app:frontier_models}. 

\subsection{Evaluation Protocol}

For each dataset, we sample 1,000 examples from the evaluation split
and repeat experiments with three random seeds (42, 43, 44) to account
for subset and perturbation variance. Models are evaluated with greedy
decoding, and accuracy is the primary metric. We report mean accuracy
across seeds. Standard deviations are small, typically below one
percentage point. Main-text results focus on severity $r=0.1$, with
all severity levels reported in Appendix~\ref{app:full_robustness}.
For GSM8K, we extract the final numeric answer following the standard
\texttt{\#\#\#\#} format. In addition to accuracy, we record tokenization
statistics and attention diagnostics to connect prediction-level
degradation with internal processing changes.

\section{How Robust Are LLMs to Lexical Perturbations?}
\label{sec:robustness_results}

We first examine whether lexical perturbations produce systematic reasoning
failures, and whether these failures are specific to character-level
corruption rather than prompt length. Table~\ref{tab:robustness_r01}
reports accuracy at severity $r=0.1$, while
Figure~\ref{fig:section4_tripanel_summary} summarizes trends across all
severity levels. Full results are provided in
Appendix~\ref{app:full_robustness}.

\paragraph{Character-level corruption degrades reasoning across models.}
As shown in Table~\ref{tab:robustness_r01}, keyboard noise and typoswap
consistently reduce accuracy across models and tasks, while filler insertion
has minimal effect. For example, Qwen3.5-9B drops from 0.845 to 0.706 on
GSM8K under keyboard noise and to 0.569 under typoswap, whereas filler
insertion leaves accuracy nearly unchanged (0.840). This contrast provides
the first evidence that the failures are tied to lexical-form corruption
rather than prompt length alone.

\paragraph{Degradation scales with perturbation severity.}
Figure~\ref{fig:section4_tripanel_summary} (left) shows that average
accuracy decreases monotonically as corruption severity increases from
$r=0.05$ to $r=0.30$ under both keyboard and typoswap perturbations.
At $r=0.30$, average accuracy drops below 0.40 for both character-level
perturbations, compared to 0.74 on clean prompts. Filler insertion remains
near the clean baseline across severities.

\paragraph{Multi-step reasoning is most vulnerable.}
Figure~\ref{fig:section4_tripanel_summary} (middle, right) shows substantial
task variation. GSM8K exhibits the steepest degradation, falling from 0.619
to 0.191 under keyboard noise and from 0.549 to 0.112 under typoswap as
severity increases from $r=0.05$ to $r=0.30$. BoolQ is comparatively robust,
suggesting that tasks requiring precise numerical interpretation and
multi-step reasoning are more sensitive to lexical corruption. 
This ordering is not explained by the amount of attention diversion: on
Llama-3.1-8B, GSM8K sustains a lower diversion ratio than BoolQ yet loses
several times as much accuracy (Appendix~\ref{app:layerwise_pertask}). What
differs is whether the displaced evidence is recoverable from elsewhere in
the prompt, which Section~\ref{sec:task_resilience} quantifies.

\begin{table*}[t]
\centering
\small
\caption{\textbf{Accuracy under lexical perturbations at severity $r=0.1$}
($n=1{,}000$). Character-level perturbations (KB, TS) consistently reduce
accuracy, whereas filler insertion (Fil) has minimal effect. Cells are shaded
by degradation relative to clean: \colorbox{drop!45}{$\ge$0.15},
\colorbox{drop!22}{0.05--0.15}.}
\label{tab:robustness_r01}
\setlength{\tabcolsep}{3.2pt}
\renewcommand{\arraystretch}{1.05}
\begin{tabular}{l cccc cccc cccc cccc}
\toprule
& \multicolumn{4}{c}{\textbf{BoolQ}} & \multicolumn{4}{c}{\textbf{PIQA}}
& \multicolumn{4}{c}{\textbf{HellaSwag}} & \multicolumn{4}{c}{\textbf{GSM8K}} \\
\cmidrule(lr){2-5}\cmidrule(lr){6-9}\cmidrule(lr){10-13}\cmidrule(lr){14-17}
\textbf{Model} & Cln & KB & TS & Fil & Cln & KB & TS & Fil
& Cln & KB & TS & Fil & Cln & KB & TS & Fil \\
\midrule
Gemma-2-9B-IT
& .875 & \cellcolor{drop!22}.809 & \cellcolor{drop!22}.826 & .878
& .843 & \cellcolor{drop!45}.673 & \cellcolor{drop!45}.645 & .827
& .693 & \cellcolor{drop!45}.507 & \cellcolor{drop!45}.469 & .697
& .804 & \cellcolor{drop!22}.729 & \cellcolor{drop!45}.586 & .802 \\
Llama-3.1-8B-Instr.
& .834 & \cellcolor{drop!22}.723 & \cellcolor{drop!22}.752 & .835
& .758 & \cellcolor{drop!45}.586 & \cellcolor{drop!22}.630 & .753
& .493 & \cellcolor{drop!22}.373 & \cellcolor{drop!45}.354 & .494
& .698 & \cellcolor{drop!45}.518 & \cellcolor{drop!45}.420 & .709 \\
Mistral-7B-Instr.-v0.3
& .847 & \cellcolor{drop!22}.731 & \cellcolor{drop!22}.758 & .848
& .693 & \cellcolor{drop!45}.549 & \cellcolor{drop!45}.541 & .695
& .413 & \cellcolor{drop!22}.347 & \cellcolor{drop!22}.336 & .408
& .397 & \cellcolor{drop!22}.276 & \cellcolor{drop!45}.201 & .378 \\
Qwen3.5-9B
& .878 & \cellcolor{drop!22}.803 & \cellcolor{drop!22}.812 & .870
& .892 & \cellcolor{drop!45}.725 & \cellcolor{drop!45}.721 & .884
& .863 & \cellcolor{drop!45}.693 & \cellcolor{drop!45}.708 & .873
& .845 & \cellcolor{drop!22}.706 & \cellcolor{drop!45}.569 & .840 \\
\bottomrule
\end{tabular}
\end{table*}

\begin{figure*}[t]
\centering
\includegraphics[width=\textwidth]{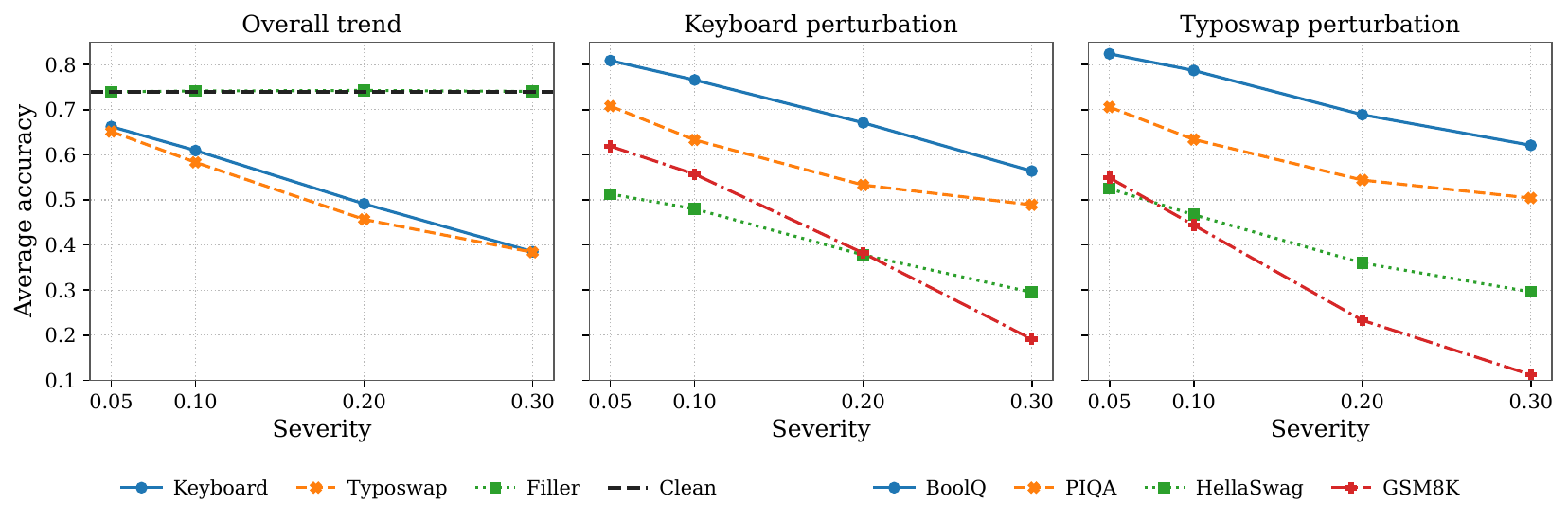}
\caption{\textbf{Impact of lexical perturbations on reasoning accuracy.}
Left: average accuracy across severity. Middle and right: task-wise
performance under keyboard noise and typoswap. Character-level
perturbations degrade monotonically with severity, while filler stays
close to clean.}
\label{fig:section4_tripanel_summary}
\end{figure*}

\begin{table}[t]
\centering
\footnotesize
\setlength{\tabcolsep}{3pt}
\caption{\textbf{Cross-model comparison at $r=0.1$}, averaged over the four
benchmarks. GPT-5.4 Thinking sustains the largest keyboard degradation of
any configuration despite near-identical clean accuracy to Standard.
Per-task results are in Appendix~\ref{app:frontier_models}.}
\label{tab:frontier_comparison}
\begin{tabular}{lccc}
\toprule
\textbf{Model} & \textbf{Clean} & \textbf{Keyboard} & \textbf{Typoswap} \\
\midrule
GPT-5.4 (Standard) & .908 & .815 \tiny(-.093) & .858 \tiny(-.050) \\
GPT-5.4 (Thinking) & .905 & .758 \tiny(-.147) & .818 \tiny(-.087) \\
GPT-4o             & .878 & .798 \tiny(-.080) & .818 \tiny(-.060) \\
\midrule
Qwen3.5-9B         & .869 & .732 \tiny(-.137) & .702 \tiny(-.167) \\
Gemma-2-9B         & .804 & .680 \tiny(-.124) & .632 \tiny(-.172) \\
Llama-3.1-8B       & .696 & .550 \tiny(-.146) & .539 \tiny(-.157) \\
Mistral-7B         & .588 & .476 \tiny(-.112) & .459 \tiny(-.129) \\
\bottomrule
\end{tabular}
\end{table}

\paragraph{The pattern extends beyond the main setting.}
The same pattern holds beyond 7B--9B English models
(Appendix~\ref{app:scale_lang}). At 70B scale, fragility persists but is
attenuated: GSM8K drops 13 points under typoswap, while BoolQ is stable. On
MATH algebra, typoswap again produces the largest drop. On Chinese CMath,
keyboard perturbation causes a 30-point drop while filler remains harmless.
Frontier models show the same pattern, and reasoning does not absorb it
(Table~\ref{tab:frontier_comparison}). GPT-5.4's thinking mode sustains the
largest keyboard degradation of any configuration we evaluate ($-0.147$),
1.6$\times$ that of the same model with reasoning disabled, despite
near-identical clean accuracy. Per-task results are in
Appendix~\ref{app:frontier_models}.

These results establish a robust empirical pattern: character-level
perturbations degrade reasoning, while filler insertion largely does not.
This raises the central mechanistic question: why does corruption of lexical
form matter more than added prompt length? We answer this question next by
analyzing tokenization and attention.

% % -------------------------------Section 5：What Causes Lexical Fragility in LLM Reasoning? ------------------------------------%

\section{What Causes Lexical Fragility in LLM Reasoning?}
\label{sec:mechanism}

Our analysis tests a three-step mechanism. Lexical corruption fragments
subword tokenization, changing the token identities the model consumes. The
fragments then attract disproportionate attention mass, producing Attention
Diversion. Fragmentation in turn drives reasoning failure beyond any effect
of prompt length. A factorial intervention (Section~\ref{sec:factorial})
then shows the second and third steps to be a coupling between attention and
token content rather than a one-directional chain.
Appendix~\ref{app:pipeline} illustrates the full pathway.

\subsection{From Fragmentation to Attention Diversion}

Lexical perturbations change the token sequence presented to the model in
two ways. Character-level perturbations substantially increase the number
of subword tokens, and this increase grows with corruption severity, while
filler insertion remains close to the clean baseline. 
Token overlap shows the same
pattern. Keyboard noise and typoswap sharply reduce overlap with the clean
prompt, while filler leaves it largely intact
(Appendix~\ref{app:tokenization_error}).

This disruption alters attention allocation during inference. We measure the
\emph{diversion ratio}, defined as the fraction of attention mass assigned to
corrupted tokens. Keyboard noise and typoswap consistently increase
corrupted-token attention across models and severities. Filler insertion
does not, staying below 0.025 throughout. The gap widens with severity. At
$r{=}0.30$, typoswap reaches 0.527 on Qwen3.5-9B, 0.299 on Gemma-2-9B,
0.211 on Mistral-7B and 0.172 on Llama-3.1-8B, against 0.448, 0.256, 0.176
and 0.144 under keyboard noise (Appendix~\ref{app:cross_tokenizer}). The
diversion ratio often exceeds the fraction of corrupted tokens in the
prompt, indicating that corrupted fragments act as attention attractors.

\begin{figure}[t]
\centering
\includegraphics[width=\linewidth]{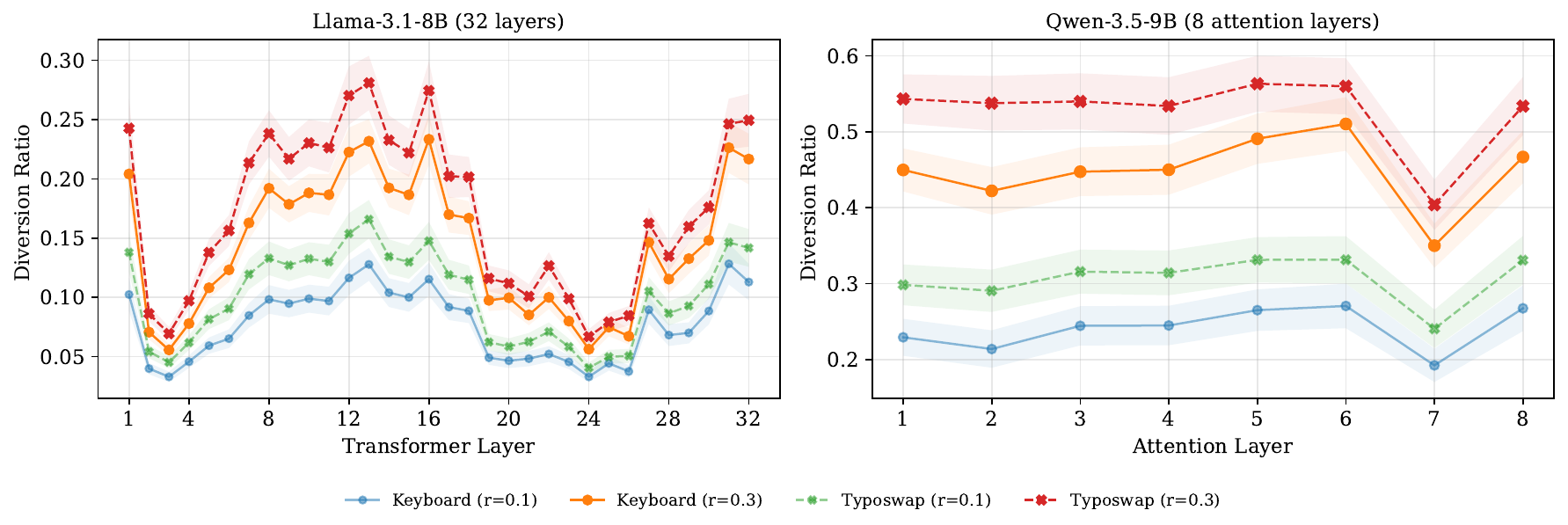}
\caption{\textbf{Layer-wise diversion ratio under lexical corruption.}
Llama-3.1-8B (top) peaks in middle and final layers, while Qwen3.5-9B
(bottom) shows stronger late-layer diversion. Averaged across four
benchmarks. Shading indicates standard error.}
\label{fig:layerwise_hijack}
\end{figure}

\paragraph{Diversion is layer-structured rather than uniform.}
To understand \emph{where} in the transformer stack diversion occurs, we
compute the diversion ratio at each layer.
Figure~\ref{fig:layerwise_hijack} shows that Attention Diversion is
structured rather than uniform. In Llama-3.1-8B, diversion forms a bimodal
pattern, peaking in middle layers and again near the final layers.
Qwen3.5-9B shows higher overall diversion, with elevated ratios toward
later attention layers. These patterns suggest that lexical corruption
affects both contextual composition and prediction-facing representations.
Per-task breakdowns in Appendix~\ref{app:layerwise_pertask} show the same
qualitative pattern.

\subsection{Fragmentation, Not Length, Drives Failure}
The preceding analyses establish that character-level perturbations
simultaneously increase fragmentation and degrade accuracy, but leave the
causal role of fragmentation ambiguous. The loss could stem from
fragmentation itself or merely from increased prompt length. We separate
the two with a controlled experiment on the same GSM8K samples
(Llama-3.1-8B, $n{=}500$, $r{=}0.1$) comparing \textbf{Group~A} (filler
insertion, which adds tokens but preserves subword structure),
\textbf{Group~B} (keyboard noise and typoswap, which disrupt token identity
while preserving semantics), and \textbf{Group~C} (neutral word padding
calibrated to match Group~B's token count). Table~\ref{tab:controlled_groups}
summarizes the results. Group~A adds more tokens than Group~B (63 vs.\ 62)
yet maintains near-clean accuracy (0.686 vs.\ 0.694), so prompt length alone
does not impair reasoning. At matched token counts, Group~B has
8$\times$ the fragmentation of Group~C (0.080 vs.\ 0.010) and lower accuracy
(0.642 vs.\ 0.662), isolating fragmentation beyond length as the driver.
The typoswap row does not follow, carrying the same fragmentation as
keyboard noise (0.079 vs.\ 0.080) yet a higher accuracy (0.682 vs.\ 0.642),
so fragmentation magnitude alone does not fix the size of the drop within a
single cell. We therefore estimate its effect across perturbation types with
the regression on the $n{=}4{,}160$ set (Appendix~\ref{app:regression})
rather than from this comparison.

Across the full set, fragmentation is the strongest univariate predictor of
a correct-to-wrong flip ($\beta{=}+0.66$, OR$=$1.93, $p{<}10^{-81}$).
Controlling for token-count change strengthens it ($\beta{=}+1.04$,
OR$=$2.84) while token-count change loses independent positive predictive
power (Appendix~\ref{app:regression}). Conditioning on the outcome shows the
same signature. Examples that flip to incorrect exhibit higher diversion
ratios, lower token overlap, and larger token-count inflation than those
that remain correct, especially at low-to-moderate corruption
(Appendix~\ref{app:tokenization_error}).

\paragraph{Both channels are load-bearing.}
Two interventions show that neither the corrupted content nor the attention
it attracts is inert. Replacing corrupted token embeddings with nearby
clean-token embeddings on GSM8K (Llama-3.1-8B, $r{=}0.1$ keyboard,
$n{=}100$) improves accuracy from 0.42 to 0.52 against a clean baseline of
0.80, recovering 26\% of the gap with only 2.95 replacements per example
(Appendix~\ref{app:embedding_intervention}). Conversely, suppressing
attention to outlier tokens by computing $s_j=\|h_j-\bar{h}\|_2$ and
applying $\tilde{A}_{ij}=A_{ij}\cdot\sigma(-\lambda s_j)$ with
$\lambda\in\{0.5,0.9\}$ consistently \emph{worsens} accuracy, with GSM8K
dropping up to $-$0.31 (Appendix~\ref{app:suppression}). Corrupted-token
attention is therefore not removable noise: the model relies on the
fragments to partially reconstruct the intended input. Recovery on one side
and harm on the other suggest the channels are not independent.

\begin{table}[t]
\centering
\footnotesize
\setlength{\tabcolsep}{4pt}
\caption{\textbf{Controlled fragmentation experiment on GSM8K}
(Llama-3.1-8B, $r{=}0.1$, $n{=}500$). At matched token count, keyboard
noise fragments more than length-matched padding and scores lower. Typoswap
does not follow, and Appendix~\ref{app:regression} estimates the effect
across perturbation types.
}
\label{tab:controlled_groups}
\begin{tabular}{lcccc}
\toprule
\textbf{Group} & \textbf{Tokens} & \textbf{Frag.} & \textbf{Overlap} & \textbf{Acc.} \\
\midrule
Clean              & 59 & 0.000 & 1.000 & 0.694 \\
A (Filler)         & 63 & 0.010 & 0.990 & 0.686 \\
C (Length-matched) & 62 & 0.010 & 0.990 & 0.662 \\
B (Keyboard)       & 62 & 0.080 & 0.920 & 0.642 \\
B (Typoswap)       & 62 & 0.079 & 0.921 & 0.682 \\
\bottomrule
\end{tabular}
\end{table}

\subsection{Attention and Token Content Are Coupled}
\label{sec:factorial}

Suppression shows that corrupted-token attention cannot be removed without
cost, but leaves open what the model would do if that attention were correct
while the content remained corrupted, or the reverse. We therefore cross the
source of the attention pattern with the source of the token embeddings.
Clean pre-softmax attention logits are injected into the corrupted forward
pass through a difflib position map on both axes, with the corrupted causal
mask applied afterwards and the softmax recomputed. Embeddings are restored
through the same map. We report recovery,
$(A_{\text{interv}}-A_{\text{corrupt}})/(A_{\text{clean}}-A_{\text{corrupt}})$,
on GSM8K at $r{=}0.1$ ($n{=}100$ per cell, bootstrap $B{=}2000$).
Implementation and validation are in Appendix~\ref{app:factorial_protocol}.

\begin{table}[t]
\centering\footnotesize
\caption{\textbf{Factorial restoration of attention and token content on
GSM8K} ($r{=}0.1$, $n{=}100$ per cell, paired cluster bootstrap $B{=}2000$).
Percentages of the clean--corrupted gap recovered, 95\% CIs below.
\colorbox{mag!30}{Grey} marks recovery and \colorbox{drop!45}{purple} harm.
Unshaded cells span zero. Neither channel alone recovers accuracy, and
clean attention over corrupted content is harmful, yet restoring both is
strongly super-additive.}
\label{tab:factorial}
\setlength{\tabcolsep}{3pt}
\renewcommand{\arraystretch}{0.95}
\begin{tabular}{@{}l cccc@{}}
\toprule
\textbf{Restored} & \textbf{Llama} & \textbf{Mistral} & \textbf{Gemma} & \textbf{Qwen}$^{\ast}$ \\
\textit{\scriptsize gap (pt)} & \scriptsize 34 & \scriptsize 24 & \scriptsize 22 & \scriptsize 14 \\
\midrule
Attention   & \hA{-41}  & $-25$ & \hA{-59}  & \hA{-164} \\
            & \cis{-88}{-8} & \cis{-87}{+10} & \cis{-164}{-3} & \cis{-625}{-64} \\
\addlinespace[2pt]
Content     & \rA{+56}  & $+25$ & $-27$ & $+21$ \\
            & \cis{+24}{+88} & \cis{-8}{+56} & \cis{-100}{+15} & \cis{-100}{+83} \\
\addlinespace[2pt]
Both        & \rA{+71}  & \rA{+54} & \rA{+45} & $-50$ \\
            & \cis{+41}{+103} & \cis{+17}{+94} & \cis{+9}{+79} & \cis{-300}{+18} \\
\midrule
Interaction & \rA{+56}  & \rA{+54} & \rA{+132} & \rA{+93} \\
            & \cis{+11}{+117} & \cis{+5}{+131} & \cis{+56}{+280} & \cis{+8}{+384} \\
\bottomrule
\end{tabular}\\[2pt]
{\scriptsize $^{\ast}$Recovery's denominator is the clean--corrupted gap,
so variance grows as the gap narrows. Qwen's 14-point gap inflates all four
intervals, and the same holds for BoolQ, which we do not decompose.}
\end{table}

Table~\ref{tab:factorial} shows that neither channel is sufficient on its
own. Reading down the first row, clean attention over corrupted content is significantly negative in
three of four models: a correctly aimed attention pattern still retrieves
the wrong token identities, so directing the model at corrupted positions
makes matters worse. Clean embeddings under a corrupted attention pattern
reach significance in only one model, which places the 26\% recovered by
embedding replacement in context as what a content-side intervention
achieves while attention remains misallocated. Restoring both together
recovers 45--71\% of the gap in the three models where the effect is
significant, with a significantly positive interaction in every model.

The channels are thus strongly super-additive. Fragmentation does not damage
token identity first and divert attention afterwards. It damages both at
once, and attention is useful only when the content it addresses is intact.
This also explains why suppressing corrupted-token attention degrades
accuracy, and why every single-channel mitigation in
Section~\ref{sec:mitigation} fails.

\subsection{What Determines Task Resilience}
\label{sec:task_resilience}

Section~\ref{sec:robustness_results} shows that diversion magnitude does
not predict task-level damage. We now test what does. 
If resilience depends on whether displaced evidence is recoverable, then
numeric content should be the vulnerable case in mathematical reasoning. A
quantity appears once, determines the answer, and cannot be inferred from
surrounding text. A corrupted content word often can.

We fit a joint logistic regression over GSM8K clean-correct records
($n{=}237$ observations, four models, typoswap at $r{=}0.1$), modelling the
correct-to-wrong flip. The numeric-token hit, the corrupted-token fraction,
and the diversion ratio enter simultaneously, with model fixed effects and
standard errors clustered by prompt (Table~\ref{tab:numeric_regression}).
Whether corruption lands on a numeral is the only significant predictor.
The flip rate is 0.77 when a numeral is corrupted and 0.33 otherwise.
Neither the corrupted-token fraction nor the diversion ratio retains
independent predictive power once prompt-level dependence is accounted for.

\begin{table}[t]
\centering\small
\caption{\textbf{Predictors of correct-to-wrong flips on GSM8K}
($n{=}237$, four models, model fixed effects, standard errors clustered by
prompt). Continuous predictors are standardized. OR is reported with its
95\% confidence interval.}
\label{tab:numeric_regression}
\setlength{\tabcolsep}{5pt}
\begin{tabular}{@{}lrl@{}}
\toprule
% \textbf{Predictor} & \textbf{OR \ci{95\% CI}{}} & \textbf{$p$} \\
\textbf{Predictor} & \textbf{OR} {\scriptsize[95\% CI]} & \textbf{$p$} \\
\midrule
Numeric-token hit & $\mathbf{12.89}$ \cis{5.06}{32.85} & $8.4\!\times\!10^{-8}$ \\
Corrupted fraction & 1.41 \cis{0.58}{3.45} & 0.45 \\
Diversion & 1.45 \cis{0.29}{7.24} & 0.65 \\
\bottomrule
\end{tabular}
\end{table}

\begin{table}[t]
\centering
\footnotesize
\setlength{\tabcolsep}{4pt}
\caption{\textbf{GSM-IC targeted fragmentation} (four models, $n{=}400$,
typoswap $r{=}0.1$, standard errors clustered by prompt). Length and
semantics are held fixed, and only the fragmented span differs. Diversion
occurs in both conditions and only one is harmful.}
\label{tab:gsmic}
\begin{tabular}{lcc}
\toprule
\textbf{Fragmented span} & \textbf{Drop (pt)} & \textbf{95\% CI} \\
\midrule
Reasoning-relevant    & 21.0 & [14.5, 28.0] \\
Irrelevant (injected) & \phantom{0}2.5 & [$-1.2$, \phantom{0}6.2] \\
\midrule
Contrast              & 18.5 & [12.2, 25.0] \\
\bottomrule
\end{tabular}
\end{table}

The regression is observational, in that which words the perturbation lands
on is left to chance. To manipulate the target directly we use GSM-IC
\citep{shi2023large}, which augments GSM8K problems with irrelevant
injected sentences. Within each example we construct variants that differ
only in \emph{what} is fragmented, holding length and semantics fixed:
either the reasoning-relevant content, or an equal-length irrelevant
injected span. Table~\ref{tab:gsmic} reports the result. Only the
reasoning-relevant condition is harmful, and the contrast between the two
is significant ($p<10^{-4}$), although diversion occurs in both.

Attention Diversion is therefore harmful in proportion to the
irreplaceability of the evidence it displaces, which explains why the
per-task diversion ordering does not track the accuracy-loss ordering
(Appendix~\ref{app:layerwise_pertask}): GSM8K suffers the largest loss
despite sustaining less diversion than BoolQ, because its evidence is the
least redundant. It also connects to the coupling result of
Section~\ref{sec:factorial}. In GSM8K the evidence is a quantity that
appears once, so displacing attention from it and directing attention to a
corrupted version of it are equally fatal, which is why restoring clean
attention over corrupted content does not help: attention correctly aimed
at a corrupted numeral still retrieves the wrong value.

\begin{table}[t]
\centering
\footnotesize
\setlength{\tabcolsep}{3pt}
\caption{\textbf{Cross-perturbation controls} (Llama-3.3-70B, English
BoolQ + GSM8K, $n=100$, separate subset from
Appendix~\ref{app:scale_lang}). Rows ordered by increasing fragmentation.
Fragmentation is shaded by magnitude (\colorbox{mag!30}{$\ge$0.08},
\colorbox{mag!14}{0.03--0.08}) and accuracy by degradation relative to clean
(\colorbox{drop!45}{$\ge$0.15}, \colorbox{drop!22}{0.05--0.15}). OCR
fragments most yet leaves accuracy intact. Typoswap alone damages GSM8K.}
\label{tab:frag_70b_english}
\begin{tabular}{lcccc}
\toprule
& \multicolumn{2}{c}{\textbf{Frag.}} & \multicolumn{2}{c}{\textbf{Acc.}} \\
\cmidrule(lr){2-3}\cmidrule(lr){4-5}
\textbf{Perturb.} & BQ & GSM & BQ & GSM \\
\midrule
Clean              & .000 & .000 & .88 & .95 \\
ASR ($r$=0.1)      & .000 & .000 & .91 & .93 \\
ASR ($r$=0.2)      & .000 & .000 & .89 & .94 \\
Keyboard ($r$=0.1) & \mB{.049} & \mB{.035} & .89 & \dB{.90} \\
Typoswap ($r$=0.1) & \mB{.062} & \mB{.048} & .89 & \dA{\textbf{.79}} \\
OCR ($r$=0.1)      & \mA{.086} & \mB{.061} & .89 & .91 \\
OCR ($r$=0.2)      & \mA{.181} & \mA{.154} & .91 & .93 \\
\bottomrule
\end{tabular}
\end{table}

\paragraph{Cross-perturbation controls refine the fragmentation account.}
The account predicts harm mainly when perturbations create rare or
out-of-distribution subword fragments.
Table~\ref{tab:frag_70b_english} tests this with ASR homophones and OCR
character confusions. 
ASR substitutions produce zero fragmentation and leave accuracy at baseline.
OCR produces the \emph{highest} fragmentation of the three perturbations yet
still leaves accuracy near baseline, because its fragments follow familiar
subword patterns.
Only typoswap, whose
fragments are out of distribution, substantially reduces GSM8K accuracy.
Fragmentation is thus necessary but not sufficient: what matters is whether
the resulting pieces are ones the model has seen. Chinese CMath shows the
same qualitative pattern (Appendix~\ref{app:fragmentation_rates}).

\paragraph{Qualitative case study.}
A GSM8K example in Appendix~\ref{app:case_study} illustrates the full
pathway: keyboard perturbation expands the prompt from 48 to 63 tokens,
redirects 67\% of attention mass to corrupted fragments, and changes the
model output from the correct answer 14 to a copied surface number 4.

Together, these analyses support a consistent mechanism. Lexical corruption
fragments tokenization, which simultaneously alters the token identities the
model reads and redirects attention onto the fragments. The factorial
intervention shows the two to be inseparable: correcting either alone leaves
accuracy at or below the corrupted baseline, while both together recover
most of the loss.
% -------------------------------Section 6------------------------------------%

\section{Can Inference-Time Strategies Mitigate Lexical Fragility?}
\label{sec:mitigation}

If lexical fragility arises from the coupled disruption of
Section~\ref{sec:factorial}, any strategy acting on one channel alone should
recover little. We investigate chain-of-thought prompting, prompt-level
repair, a 70B repair model, and a tokenizer-level defense. Because these
experiments used different subsets and templates, each is reported against
its own paired baseline and clean anchor. Recovery is a percentage of the
clean--corrupted gap within each comparison.

\paragraph{Prompt-level repair trades one failure for another.}
Table~\ref{tab:mitigation_ledger} evaluates every mitigation on a common
recovery scale. Spell-checking is the only arm ever cleanly positive,
recovering 30\% of the gap on BoolQ under keyboard noise, yet it turns
negative under typoswap on both tasks, where transposed characters yield
fluent but incorrect dictionary matches. Noise-aware prompting is negative
in all four conditions. Self-repair recovers half the gap on GSM8K under
keyboard noise but destroys BoolQ, dropping it from 0.73 to 0.13 by
rewriting passage content the question depends on. Substituting
Llama-3.3-70B-Instruct as the rewriter recovers the BoolQ keyboard
condition fully and leaves the other three unchanged, so the failure is not
a matter of rewriter capacity. Correcting the surface form does not restore
the token identities the reasoning depends on. Reasoning-time prompting
fares no better. 
% CoT lifts clean GSM8K on GPT-4o from 0.42 to 0.90, yet
% typoswap still pulls it to 0.70, and GPT-5.4's built-in reasoning mode
% \emph{amplifies} the degradation rather than absorbing it
% (Appendix~\ref{app:frontier_models}). 
CoT lifts clean GSM8K on
GPT-4o from 0.42 to 0.90, yet typoswap still pulls it to 0.70
(Appendix~\ref{app:cot}), and GPT-5.4's built-in reasoning mode
\emph{amplifies} the degradation rather than absorbing it
(Appendix~\ref{app:frontier_models}).
Attention calibration and prompt
restoration, evaluated separately, likewise show no consistent improvement
(Appendix~\ref{app:diagnostic_probes}).

\begin{table}[t]
\centering\small
\caption{\textbf{Mitigation ledger.} Recovery as a percentage of each arm's
own clean--corrupted gap. Every arm is paired against a baseline on its own
subset and prompt, so accuracies are not comparable across blocks.
\colorbox{mag!30}{Grey} marks recovery and \colorbox{drop!45}{purple} harm.
No strategy is positive in more than two of four conditions.}
\label{tab:mitigation_ledger}
\setlength{\tabcolsep}{4.5pt}
\renewcommand{\arraystretch}{1.08}
\begin{tabular}{@{}l cc cc@{}}
\toprule
& \multicolumn{2}{c}{\textbf{BoolQ}} & \multicolumn{2}{c}{\textbf{GSM8K}} \\
\cmidrule(lr){2-3}\cmidrule(lr){4-5}
\textbf{Strategy} & KB & TS & KB & TS \\
\midrule
\multicolumn{5}{@{}l}{\textit{Prompt-level repair} --- Llama-3.1-8B,
$n{=}100$, $r{=}0.1$}\\[1pt]
\quad Spell-check  & \rA{+30}  & \hB{-10}  & \rB{+13} & \hB{-29} \\
\quad Noise-aware  & \hA{-150} & \hA{-80}  & \hB{-25} & \hB{-50} \\
\quad Self-repair  & \hA{-600} & \hA{-620} & \rA{+50} & $0$ \\
\addlinespace[2pt]
\multicolumn{5}{@{}l}{\quad\scriptsize clean .83/.81, corrupted
.73/.73/.65/.67}\\
\midrule
\multicolumn{5}{@{}l}{\textit{Stronger rewriter} --- Llama-3.3-70B-Instruct}\\[1pt]
\quad 70B repair   & \rA{+114}$^{\ast}$ & $0$ & $0$ & $0$ \\
\addlinespace[2pt]
\multicolumn{5}{@{}l}{\quad\scriptsize clean .79/.69, corrupted
.72/.71/.48/.49}\\
\bottomrule
\end{tabular}\\[2pt]
{\scriptsize $^{\ast}$Post-repair accuracy (0.80) exceeds the clean anchor
(0.79), so recovery exceeds 100\%. The excess is within sampling
noise.}\\[4pt]
\begin{minipage}{\columnwidth}
\scriptsize
\textbf{Tokenizer-level defense} (4 models, GSM8K and BoolQ, typoswap
$r{=}0.1$, $n{=}100$/cell). Detection reaches F1 0.77--0.81 without clean
text, and fragmentation alone reaches 0.70--0.75. Repair restores the
original word 67--70\% of the time at a false-repair rate of 22--25\%.
Recovery is positive in 0 of 8 cells and negative in 1, with no clean-input
damage. \textbf{Oracle} (clean token IDs restored). Full recovery on all
models.
\end{minipage}
\end{table}

\subsection{A Tokenizer-Level Defense}
\label{sec:defense}

The strategies above all operate on the prompt. We now test whether the
damage can be addressed where it originates, bounding the problem from
both ends: an oracle with access to the clean text, and a deployable
defense without it. Headline numbers are given at the foot of
Table~\ref{tab:mitigation_ledger}. The protocol is in
Appendix~\ref{app:defense_protocol}.

\paragraph{An oracle ceiling.}
Replacing each corrupted word's fragmented tokens with the clean word's
exact token IDs restores accuracy to the clean baseline on all four models
(GSM8K, typoswap $r{=}0.1$, $n{=}100$). This is a by-construction ceiling
rather than a method, since reconstructing the clean token sequence
reconstructs the clean input. Its value is diagnostic: had the perturbation
degraded the model's reasoning rather than its input, supplying the correct
tokens would not have restored performance. It did, so the reasoning is
intact and the damage is confined to the token sequence.

\paragraph{A deployable defense.}
We then built a defense under the constraint that makes the result
interpretable. The detector and the repair observe only the corrupted
input, with no access to the clean text or to the perturbation mask, which
is used solely for post-hoc scoring. Detection combines rarity, surprisal,
and fragmentation. Repair uses vocabulary-constrained edit-distance
correction, with a high-precision detector variant and a conservative
variant that declines when no confident candidate exists. Thresholds were
selected on a held-out split disjoint from the evaluation subset.
Character-level fallback and greedy retokenization were implemented but not
deployed, because they rewrite every flagged word and so inherit the
detector's false-repair rate of 22--25\%, the failure mode that drives
self-repair from 0.73 to 0.13 on BoolQ.

Detection succeeds while repair fails. Fragmentation alone identifies
corrupted words at F1 0.70--0.75 with no access to the clean text, which
shows it is a reliable signature of corruption and not only a consequence
of it. Repair does not follow: vocabulary-constrained correction recovers
the original word only 67--70\% of the time, and the substitutions that are
confident but incorrect do more harm than the fragments they replace, since
a fragment at least signals anomaly whereas a fluent substitution silently
alters the semantics. Recovery is significantly positive in none of the
eight model--task cells and significantly negative in one. The defense does
not damage clean inputs, unlike self-repair, but it does not help either.

The binding constraint is informational. Numeric tokens carry the largest
effect on GSM8K failure (Section~\ref{sec:task_resilience}), yet a
corrupted numeral remains a valid numeral: it is not out of vocabulary, it
is too short to fragment, and it admits no dictionary neighbour, so no
detector operating on the corrupted input can find it and no corrector can
recover the original quantity. 
Taken with the oracle result, this locates the problem precisely. The damage
is localized at the input representation, since reconstructing the clean
token sequence restores accuracy in full. But it is irreversible without the
clean text, because the corruptions that dominate failure leave no
recoverable trace.

\paragraph{Summary.}
No strategy recovers consistently, matching the coupling of
Section~\ref{sec:factorial}: each acts on one channel while the other
remains corrupted, and the corruptions that matter most leave no
recoverable trace in the corrupted input.

\section{Related Work}
\label{sec:related_work}

Prior work established the phenomenon: corrupted fragments attract
disproportionate attention. Our factorial intervention shows that this
attention cannot be manipulated on its own, a finding that organizes three
lines of research, each locating the remedy elsewhere. Typo-attack work
locates it at the input, restoring the surface form before the model reads it
\cite{belinkov2018synthetic,pruthi2019combating,li2019textbugger,ebrahimi2018hotflip,jia2017adversarial}.
Such correction is bounded by what the corrupted input still encodes, and a
perturbed numeral stays in vocabulary with no dictionary neighbour
(Section~\ref{sec:defense}). Attention-sink work locates it in the attention
pattern, suppressing positions that absorb attention regardless of content
\cite{xiao2024efficient}. The sinks we observe are created by the
perturbation rather than fixed by the model, and suppressing them costs up to
$0.31$ on GSM8K (Appendix~\ref{app:suppression}). Subword regularization
locates it in training \cite{provilkov2020bpe}, the only remaining option
once fragmentation has occurred and the information needed to reverse it is
gone. Appendix~\ref{app:related_work} develops these comparisons.

\section{Conclusion}
This paper studies how realistic lexical perturbations disrupt LLM reasoning
across four benchmarks, four open-weight model families, and frontier
systems. Character-level noise fragments subword tokenization, which alters
the token identities the model reads and redirects attention onto the
fragments at once, while length-matched filler stays benign. Controlled
experiments, per-example regression, and representation-level interventions
identify fragmentation rather than prompt length as the driver, and a
factorial intervention shows the two channels to be coupled rather than
sequential. This coupling explains why no inference-time strategy recovers
consistently, and why harm tracks what the corruption lands on rather than
how much attention it diverts. The damage is localized at the input
representation yet irreversible without the clean text, so robustness must be
built into the representation before fragmentation occurs rather than
repaired afterwards. Character- and byte-aware fallback representations,
dynamic retokenization, and training-time alignment all act at that point,
and are the directions we see as most promising.

\section*{Limitations}
We note several boundaries. First, the main analyses focus on English with a
QWERTY layout, though we extend the protocol to Chinese (CMath) and OCR/ASR
controls. Broader coverage of languages and input methods remains open.
Second, the perturbations are synthetic, approximating common noisy-input
phenomena rather than sampled from user data. Third, all benchmarks are
standard-length single-turn tasks. Long-context and agentic settings
introduce position and interaction effects that would need separate control.
Fourth, the factorial intervention is run on GSM8K at a single severity,
since BoolQ's narrow gap makes the recovery ratio ill-conditioned, so the
coupling result rests on one task. Fifth, our interventions are diagnostic
and do not identify the internal circuits responsible for failure. Finally,
we study open-weight models in the 7B--70B range and frontier
configurations, and other recipes or scales may differ.

\section*{Ethical Considerations}
This work does not involve human subjects, sensitive personal data, or any
proprietary datasets. All datasets used are publicly available and commonly
used in prior research. We have taken care to ensure that our methods and
results do not raise safety, privacy, or fairness concerns.

\section*{AI Assistance Disclosure}
AI assistance was used only for language polishing and was not used to
generate experimental results or analyses. The authors verified all
scientific claims, experiments, analyses, and final text.

\clearpage

% -------------------------------Section 9：Limitations ------------------------------------%

\bibliography{reference}

\newpage

\appendix
\twocolumn[%
  \begin{center}
    {\LARGE\textbf{Appendix}}
  \end{center}
  \vspace{1.5em}
]

\section{Frontier LLM Robustness Evaluation}
\label{app:frontier_models}

To examine whether the lexical robustness patterns observed in open-weight
models extend to proprietary frontier systems, we evaluate GPT-4o
\cite{gpt4o}, GPT-5.4 with \texttt{reasoning\_effort=none} (``Standard''),
and GPT-5.4 with \texttt{reasoning\_effort=high} (``Thinking''). Without
access to internal representations such as tokenization states or attention
weights, these models are evaluated only at the prediction level. The
protocol follows the perturbation procedure and prompt format of the main
experiments on a subset of 100 examples per dataset, with two representative
perturbation types (keyboard noise and typoswap) at severity $r=0.1$.

\subsection{Chain-of-Thought Prompting}
\label{app:cot}
Table~\ref{tab:cot_results} evaluates direct versus CoT prompting on GPT-4o
($r{=}0.1$, $n{=}100$). CoT lifts clean GSM8K from 0.42 to 0.90, confirming
that the model's reasoning capability is intact, yet typoswap still pulls
it to 0.70. On the other three benchmarks CoT provides no consistent
advantage over direct prompting, indicating that stronger reasoning does
not repair corrupted representations. The following sections extend this
to built-in reasoning.

\begin{table}[!ht]
\centering
\small
\caption{\textbf{Direct vs.\ CoT prompting} (GPT-4o, $r{=}0.1$, $n{=}100$).
CoT lifts clean GSM8K from 0.42 to 0.90 but leaves the corruption gap
intact. Cells are shaded by degradation relative to each mode's own clean
accuracy: \colorbox{drop!45}{$\ge$0.15}, \colorbox{drop!22}{0.05--0.15}.}
\label{tab:cot_results}
\setlength{\tabcolsep}{4pt}
\begin{tabular}{@{}lcc cc cc@{}}
\toprule
& \multicolumn{2}{c}{\textbf{Clean}} & \multicolumn{2}{c}{\textbf{Keyboard}} & \multicolumn{2}{c}{\textbf{Typoswap}} \\
\cmidrule(lr){2-3}\cmidrule(lr){4-5}\cmidrule(lr){6-7}
\textbf{Task} & Direct & CoT & Direct & CoT & Direct & CoT \\
\midrule
BoolQ     & .90 & .87 & \dB{.82} & .84 & .86 & .88 \\
PIQA      & .91 & .92 & \dA{.76} & \dB{.85} & .92 & .90 \\
HellaSwag & .86 & .81 & \dB{.80} & \dB{.74} & \dB{.80} & .81 \\
GSM8K     & .42 & \textbf{.90} & .50 & .86 & .40 & \dA{\textbf{.70}} \\
\bottomrule
\end{tabular}
\end{table}

\subsection{GPT-4o Results}

Table~\ref{tab:frontier_results} presents results for GPT-4o.
GPT-4o achieves substantially higher clean accuracy than the open-weight models
across all four benchmarks, yet still exhibits consistent performance degradation
under character-level perturbations.
Under keyboard noise, the accuracy drops range from 4 to 14 points depending on the task,
while typoswap perturbations produce larger degradation on GSM8K ($-$18 points)
but minimal impact on PIQA.

\begin{table}[!ht]
\centering
\small
\caption{\textbf{GPT-4o robustness under lexical perturbations} ($n=100$,
$r=0.1$). Drop relative to clean in parentheses. Shading:
\colorbox{drop!45}{$\ge$0.15}, \colorbox{drop!22}{0.05--0.15}.}
\label{tab:frontier_results}
\setlength{\tabcolsep}{4pt}
\begin{tabular}{@{}lccc@{}}
\toprule
\textbf{Task} & \textbf{Cln} & \textbf{Keyboard} & \textbf{Typoswap} \\
\midrule
BoolQ     & .90 & \dB{.82 \tiny(-.08)} & \dB{.85 \tiny(-.05)} \\
PIQA      & .92 & \dB{.78 \tiny(-.14)} & .92 \tiny(+.00) \\
HellaSwag & .84 & \dB{.78 \tiny(-.06)} & .83 \tiny(-.01) \\
GSM8K     & .85 & .81 \tiny(-.04) & \dA{.67 \tiny(-.18)} \\
\midrule
\textit{Avg.} & \textit{.878} & \textit{.798 \tiny(-.080)} & \textit{.818 \tiny(-.060)} \\
\bottomrule
\end{tabular}
\end{table}

\subsection{GPT-5.4 Standard and Thinking Results}

The two GPT-5.4 configurations use the same underlying model through the
OpenAI API and differ only in \texttt{reasoning\_effort}. Standard mode
(\texttt{reasoning\_effort=none}) disables chain-of-thought reasoning, while
thinking mode (\texttt{reasoning\_effort=high}) enables extended internal
reasoning before producing a response, so the comparison directly tests
whether built-in reasoning mitigates lexical fragility.

Table~\ref{tab:gpt54_results} presents the results. Standard mode achieves
the highest clean accuracy among all frontier models (0.908 average) and
degrades by $-$0.093 under keyboard noise, comparable to GPT-4o ($-$0.080).
Critically, \textbf{thinking mode amplifies lexical fragility rather than
mitigating it.} Its average keyboard degradation of $-$0.147 is roughly
1.6$\times$ standard mode and 1.8$\times$ GPT-4o, with the most severe drops
on HellaSwag ($-$0.23 under keyboard noise) and GSM8K ($-$0.23 under
typoswap), both multi-step reasoning tasks. This extends the CoT analysis in
Appendix~\ref{app:cot}. Reasoning-time strategies, whether explicit CoT
prompting (Table~\ref{tab:cot_results}) or built-in model reasoning, cannot
compensate for input-level tokenization disruption, and the extended
reasoning chain appears to \emph{propagate} early tokenization errors
through more computation steps rather than correcting them.

\begin{table}[!ht]
\centering
\small
\caption{\textbf{GPT-5.4 robustness under lexical perturbations} ($n=100$,
$r=0.1$). Drop relative to each mode's own clean accuracy in parentheses.
Thinking mode amplifies fragility, with the largest average keyboard drop
($-$0.147) among all frontier configurations. Shading:
\colorbox{drop!45}{$\ge$0.15}, \colorbox{drop!22}{0.05--0.15}.}
\label{tab:gpt54_results}
\setlength{\tabcolsep}{4pt}
\begin{tabular}{@{}lccc@{}}
\toprule
\textbf{Task} & \textbf{Cln} & \textbf{Keyboard} & \textbf{Typoswap} \\
\midrule
\multicolumn{4}{@{}l}{\textit{GPT-5.4 (Standard)}}\\[1pt]
BoolQ     & .88 & \dB{.77 \tiny(-.11)} & .87 \tiny(-.01) \\
PIQA      & .95 & \dA{.79 \tiny(-.16)} & \dB{.88 \tiny(-.07)} \\
HellaSwag & .83 & \dB{.77 \tiny(-.06)} & .86 \tiny(+.03) \\
GSM8K     & .97 & .93 \tiny(-.04) & \dA{.82 \tiny(-.15)} \\
\addlinespace[1pt]
\textit{Avg.} & \textit{.908} & \textit{.815 \tiny(-.093)} & \textit{.858 \tiny(-.050)} \\
\midrule
\multicolumn{4}{@{}l}{\textit{GPT-5.4 (Thinking)}}\\[1pt]
BoolQ     & .89 & \dA{.70 \tiny(-.19)} & .86 \tiny(-.03) \\
PIQA      & .91 & \dB{.83 \tiny(-.08)} & .90 \tiny(-.01) \\
HellaSwag & .86 & \dA{.63 \tiny(-.23)} & \dB{.78 \tiny(-.08)} \\
GSM8K     & .96 & \dB{.87 \tiny(-.09)} & \dA{.73 \tiny(-.23)} \\
\addlinespace[1pt]
\textit{Avg.} & \textit{.905} & \textit{.758 \tiny(-.147)} & \textit{.818 \tiny(-.087)} \\
\bottomrule
\end{tabular}
\end{table}

\subsection{Cross-Model Comparison}

\begin{figure}[!ht]
\centering
\includegraphics[width=\linewidth]{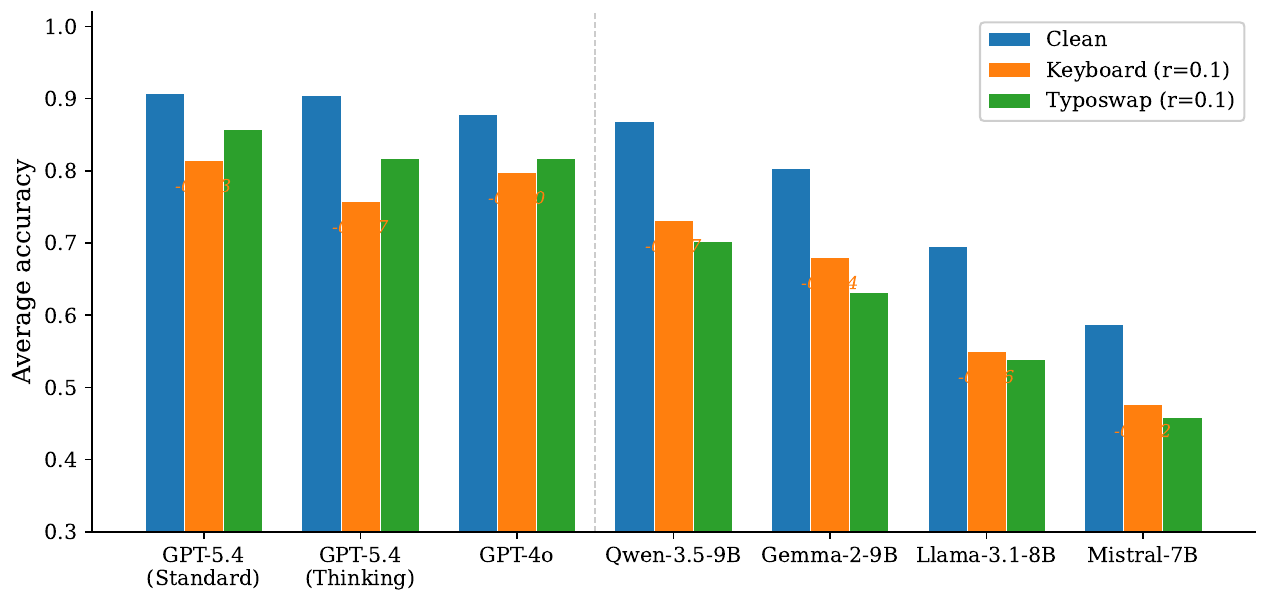}
\caption{\textbf{Cross-model comparison under lexical corruption at $r=0.1$.}
GPT-5.4 Standard achieves the highest clean accuracy.
GPT-5.4 Thinking shows the largest degradation among frontier models ($-$0.147 under keyboard noise),
confirming that reasoning amplifies rather than mitigates lexical fragility.
Numbers indicate the absolute drop under keyboard noise relative to clean prompts.}
\label{fig:frontier_comparison}
\end{figure}

\begin{figure}[!ht]
\centering
\includegraphics[width=\linewidth]{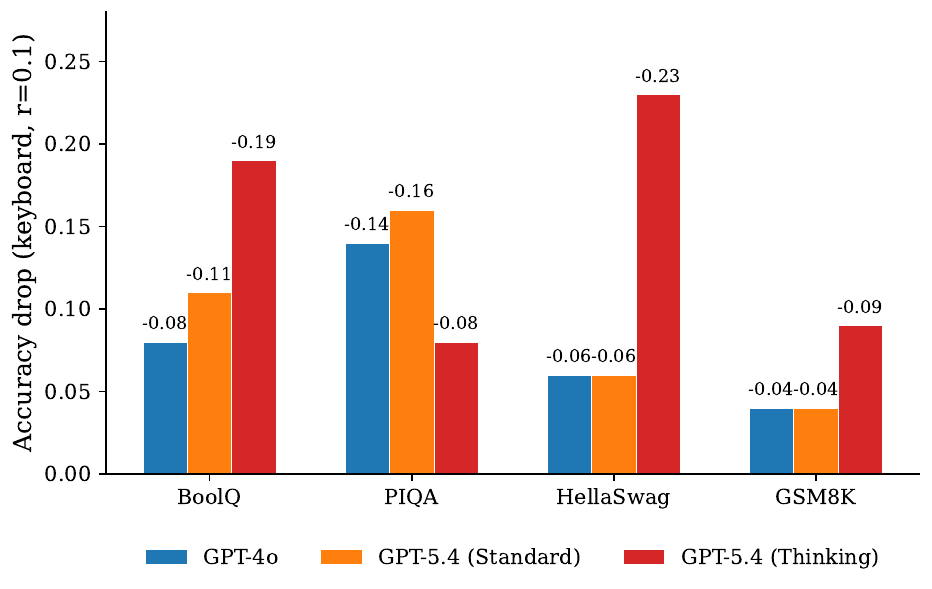}
\caption{\textbf{Task-level accuracy drop under keyboard noise at $r=0.1$ for frontier models.}
GPT-5.4 Thinking exhibits the largest degradation on HellaSwag ($-$0.23) and BoolQ ($-$0.19),
while GPT-5.4 Standard is most robust on GSM8K ($-$0.04).
Error patterns differ across reasoning modes despite near-identical clean performance.}
\label{fig:frontier_task_breakdown}
\end{figure}

% Table~\ref{tab:frontier_comparison} and Figure~\ref{fig:frontier_comparison}
% compare all frontier configurations with the open-weight models.
Table~\ref{tab:frontier_comparison} in the main text and
Figures~\ref{fig:frontier_comparison}--\ref{fig:frontier_task_breakdown}
compare all frontier configurations with the open-weight models.
Three observations emerge.
First, GPT-5.4 Standard achieves the highest clean accuracy (0.908) but its
corruption robustness does not proportionally improve, suggesting that
raw capability gains do not resolve tokenizer-level fragility.
Second, GPT-5.4 Thinking, despite near-identical clean accuracy to
standard mode, exhibits the \emph{largest} keyboard degradation ($-$0.147)
among all frontier systems, confirming that reasoning amplifies rather
than compensates for input-level disruption.
Third, the task-sensitivity pattern is preserved across all frontier models:
GSM8K remains most vulnerable, and BoolQ is most robust.

%#######################################################################

\section{Full Robustness Results at All Severity Levels}
\label{app:full_robustness}

Table~\ref{tab:robustness-all} presents detailed accuracy results
at severity levels $r \in \{0.05, 0.20, 0.30\}$,
complementing the $r=0.1$ results in Table~\ref{tab:robustness_r01}
in the main text.

\begin{table*}[!ht]
\centering
\footnotesize
\caption{\textbf{Accuracy at severity $r \in \{0.05, 0.20, 0.30\}$}
($n=1{,}000$). Shading marks degradation relative to clean:
\colorbox{drop!55}{$\ge$0.30}, \colorbox{drop!35}{0.15--0.30},
\colorbox{drop!18}{0.05--0.15}. Three levels are used here because the range
is wider than in the main text. No Fil cell reaches the 0.05 threshold at
any severity.}
\label{tab:robustness-all}
\setlength{\tabcolsep}{4pt}
\renewcommand{\arraystretch}{1.05}
\begin{tabular}{llcccc|cccc|cccc}
\toprule
& & \multicolumn{4}{c|}{$r=0.05$} & \multicolumn{4}{c|}{$r=0.20$} & \multicolumn{4}{c}{$r=0.30$} \\
\textbf{Model} & \textbf{Task} & \textbf{Cln} & \textbf{KB} & \textbf{TS} & \textbf{Fil} & \textbf{Cln} & \textbf{KB} & \textbf{TS} & \textbf{Fil} & \textbf{Cln} & \textbf{KB} & \textbf{TS} & \textbf{Fil} \\
\midrule
\multirow{4}{*}{Gemma-2-9B-IT}
 & BoolQ     & .875 & .830 & .857 & .875 & .875 & \dL{.737} & \dL{.787} & .878 & .875 & \dM{.631} & \dM{.686} & .874 \\
 & PIQA      & .843 & \dL{.736} & \dL{.725} & .833 & .843 & \dM{.584} & \dM{.566} & .827 & .843 & \dH{.504} & \dH{.495} & .828 \\
 & HellaSwag & .693 & \dM{.520} & \dM{.528} & .686 & .693 & \dH{.386} & \dH{.383} & .687 & .693 & \dH{.323} & \dH{.320} & .685 \\
 & GSM8K     & .804 & .775 & \dL{.692} & .807 & .804 & \dM{.575} & \dH{.349} & .797 & .804 & \dH{.351} & \dH{.179} & .809 \\
\midrule
\multirow{4}{*}{Llama-3.1-8B}
 & BoolQ     & .834 & \dL{.776} & .809 & .839 & .834 & \dM{.622} & \dM{.672} & .834 & .834 & \dH{.518} & \dM{.622} & .832 \\
 & PIQA      & .758 & \dL{.653} & \dL{.677} & .761 & .758 & \dM{.482} & \dM{.527} & .755 & .758 & \dM{.505} & \dM{.502} & .753 \\
 & HellaSwag & .493 & \dL{.398} & \dL{.398} & .502 & .493 & \dM{.289} & \dM{.281} & .493 & .493 & \dM{.233} & \dM{.246} & .489 \\
 & GSM8K     & .698 & \dL{.614} & \dM{.542} & .697 & .698 & \dH{.341} & \dH{.167} & .697 & .698 & \dH{.160} & \dH{.071} & .695 \\
\midrule
\multirow{4}{*}{Mistral-7B}
 & BoolQ     & .847 & \dL{.785} & \dL{.795} & .848 & .847 & \dM{.638} & \dM{.613} & .837 & .847 & \dM{.557} & \dM{.579} & .843 \\
 & PIQA      & .693 & \dL{.638} & \dL{.620} & .692 & .693 & \dM{.459} & \dM{.479} & .697 & .693 & \dM{.454} & \dM{.479} & .704 \\
 & HellaSwag & .413 & .363 & .367 & .413 & .413 & \dL{.297} & \dL{.267} & .414 & .413 & \dL{.263} & \dM{.250} & .413 \\
 & GSM8K     & .397 & \dL{.329} & \dL{.271} & .414 & .397 & \dM{.151} & \dH{.078} & .408 & .397 & \dH{.062} & \dH{.035} & .419 \\
\midrule
\multirow{4}{*}{Qwen3.5-9B}
 & BoolQ     & .878 & .845 & .837 & .877 & .878 & \dM{.688} & \dM{.685} & .875 & .878 & \dH{.549} & \dM{.598} & .873 \\
 & PIQA      & .892 & \dL{.805} & \dL{.801} & .893 & .892 & \dM{.608} & \dM{.604} & .881 & .892 & \dH{.494} & \dH{.541} & .889 \\
 & HellaSwag & .863 & \dL{.770} & \dL{.808} & .857 & .863 & \dH{.539} & \dH{.507} & .852 & .863 & \dH{.360} & \dH{.370} & .860 \\
 & GSM8K     & .845 & \dL{.760} & \dM{.692} & .836 & .845 & \dH{.461} & \dH{.338} & .842 & .845 & \dH{.191} & \dH{.163} & .840 \\
\bottomrule
\end{tabular}
\end{table*}

%######################################################################

\section{Layer-wise diversion ratio by Task}
\label{app:layerwise_pertask}

\begin{figure*}[!ht]
\centering
\includegraphics[width=0.75\textwidth]{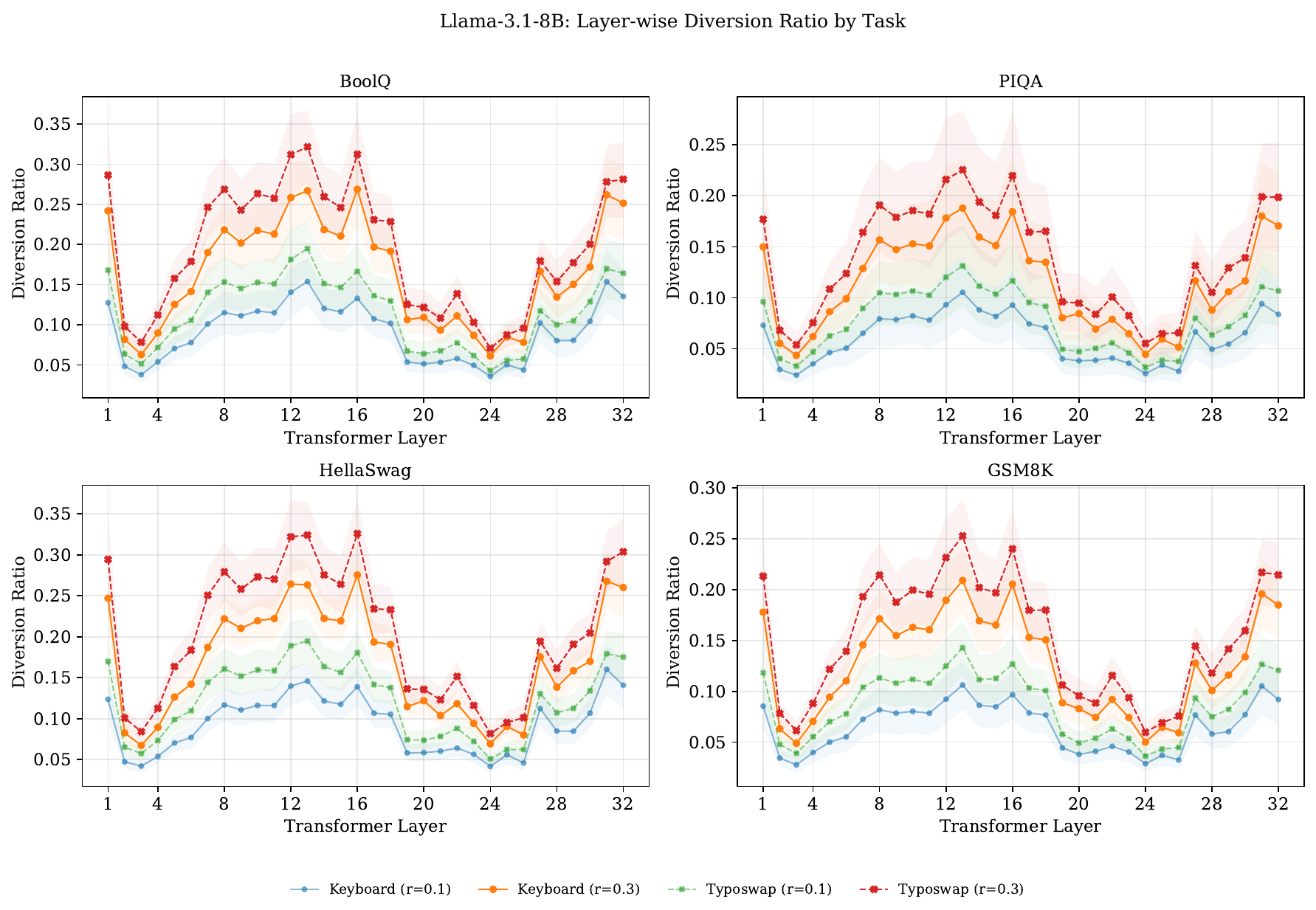}
\caption{\textbf{Layer-wise diversion ratio by task for Llama-3.1-8B.}
The bimodal pattern with peaks at middle layers (11--15) and final layers (30--32) is consistent across all four reasoning benchmarks.
BoolQ and HellaSwag show the strongest diversion, while GSM8K exhibits lower absolute values but the same structural pattern.}
\label{fig:layerwise_pertask}
\end{figure*}

Figure~\ref{fig:layerwise_pertask} presents the per-task layer-wise diversion
ratios for Llama-3.1-8B. The bimodal diversion pattern observed in the
cross-task average (Figure~\ref{fig:layerwise_hijack}) is consistently
reproduced across all four reasoning benchmarks. The task ordering is
model-dependent at the top. HellaSwag and BoolQ show the most pronounced
peaks, while GSM8K and PIQA are consistently lower. GSM8K therefore sustains
less diversion than BoolQ despite losing far more accuracy, which is the
dissociation analysed in Section~\ref{sec:task_resilience}. Per-model
orderings for the other three models are consistent and omitted for space.
These results confirm that the layer-wise distribution of Attention Diversion
is a robust architectural property rather than a task-specific artifact.

%######################################################################

\section{Generalization Across Scale, Difficulty, and Language}
\label{app:scale_lang}

To verify that the patterns observed on 7--9B open-weight models
generalize, we evaluated Llama-3.3-70B-Instruct on three additional
settings at $r=0.1$ ($n=100$). 
Section~\ref{sec:robustness_results} summarizes these results. This appendix
provides per-task details and qualitative observations.

\paragraph{70B-scale (English).}
Llama-3.3-70B was queried via the Together AI API in greedy decoding
mode. The fragility pattern is preserved: GSM8K drops 13 points under
typoswap, while BoolQ remains stable.  Keyboard noise has near-zero
effect at this scale ($-$0.01 on both tasks), while typoswap remains
substantial ($-$0.13 on GSM8K), indicating that scale partially
mitigates the milder character-substitution perturbations but does
not eliminate the more aggressive transposition pattern.

\paragraph{Harder mathematics (MATH algebra).}
We evaluated Llama-3.3-70B on the algebra subset of the MATH benchmark
\cite{hendrycks2021math} at $r=0.1$ ($n=100$).  Clean accuracy is
0.68, substantially below GSM8K (0.92), reflecting the higher
difficulty.  The same qualitative ordering holds: typoswap produces
the largest drop ($-$0.08, to 0.60), keyboard noise produces a
moderate drop ($-$0.06, to 0.62), filler is nearly harmless
($-$0.04, to 0.64).  
Extremely difficult benchmarks such as AIME or HMMT can confound this signal
when clean accuracy approaches the floor. MATH algebra provides a
non-saturated harder regime in which the fragmentation effect remains
detectable.

\paragraph{Non-Latin script (Chinese, CMath).}
We adapted the perturbation rules to operate on Chinese characters. Keyboard
perturbation replaces a character with a near-Unicode-codepoint neighbor,
where Unicode adjacency serves as a structural analog of QWERTY adjacency.
Typoswap swaps two adjacent characters. Filler inserts Chinese disfluency
phrases (transliterated \textit{``e''} or \textit{``en''}, the Chinese
equivalents of \textit{``um''} or \textit{``uh''}) at clause boundaries. On
the CMath elementary-school math benchmark, Llama-3.3-70B achieves 0.94 /
0.64 / 0.81 / 0.95 under clean / keyboard / typoswap / filler at $r=0.1$
($n=100$). The keyboard drop of 30 points reproduces the English pattern,
filler is harmless, and typoswap is intermediate. The mechanism is not
specific to Latin scripts.

%##################################################################

\section{Failure Pathway Overview}
\label{app:pipeline}

\begin{figure*}[!ht]
\centering
\includegraphics[width=\textwidth,trim=0 120 0 0,clip]{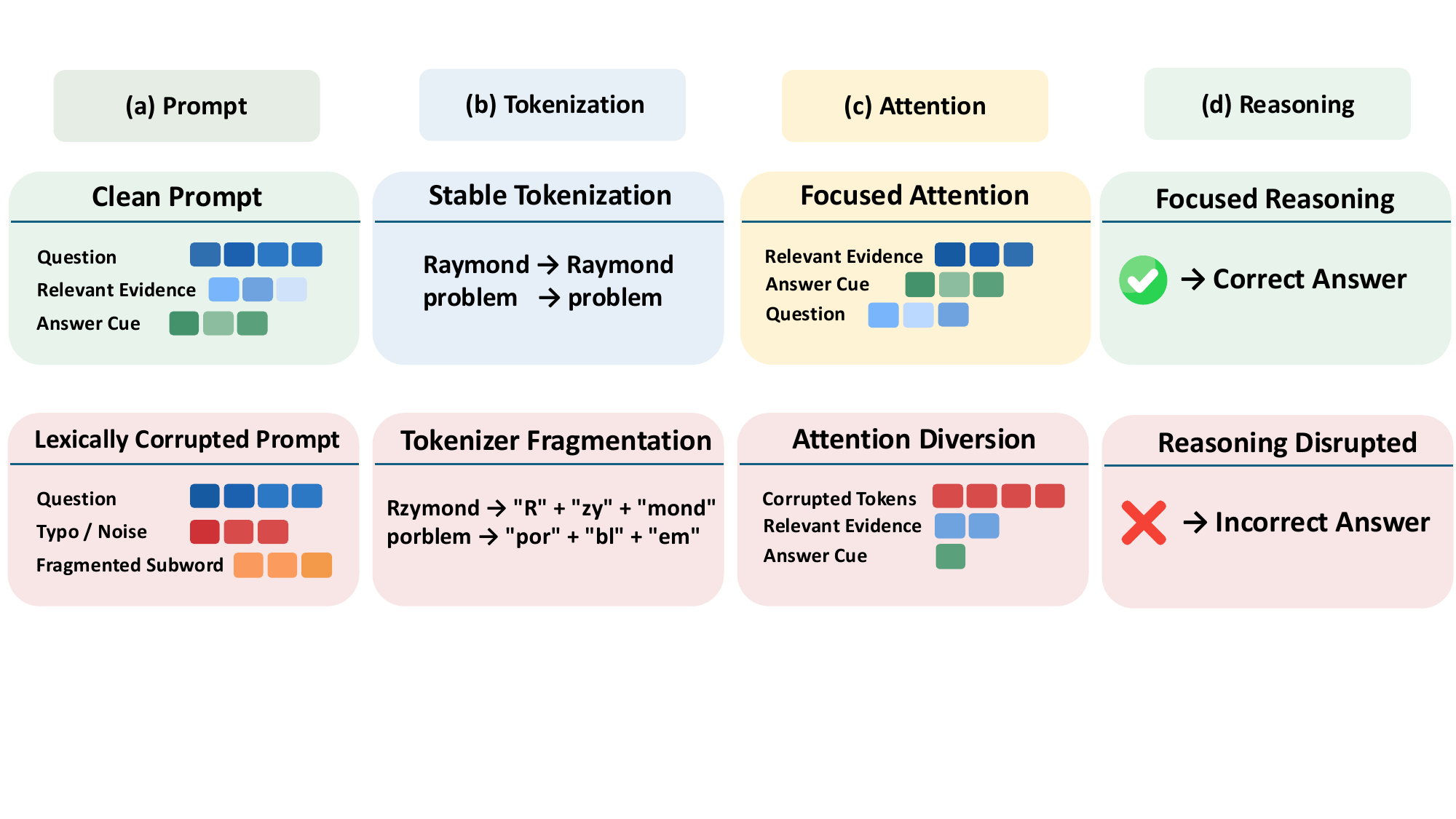}
\caption{\textbf{Complete failure pathway under lexical corruption.}
Top: clean input proceeds through stable tokenization, focused attention,
and correct reasoning.
Bottom: lexical corruption fragments tokens into novel subword pieces,
the fragments absorb a disproportionate share of attention,
and reasoning fails.}
\label{fig:pipeline_overview}
\end{figure*}

Figure~\ref{fig:pipeline_overview} illustrates the complete
processing pathway under clean and corrupted inputs.
The four stages of prompt composition, tokenization,
attention allocation, and reasoning are contrasted
to show how lexical corruption cascades through the model:
corrupted characters trigger tokenizer fragmentation~(b),
with attention mass reallocating toward noise tokens~(c),
and reasoning failing in the output stage~(d).

%###########################################################################

\section{Tokenization Disruption and Error-Conditioned Evidence}
\label{app:tokenization_error}

\begin{figure}[!ht]
\centering
\includegraphics[width=\linewidth]{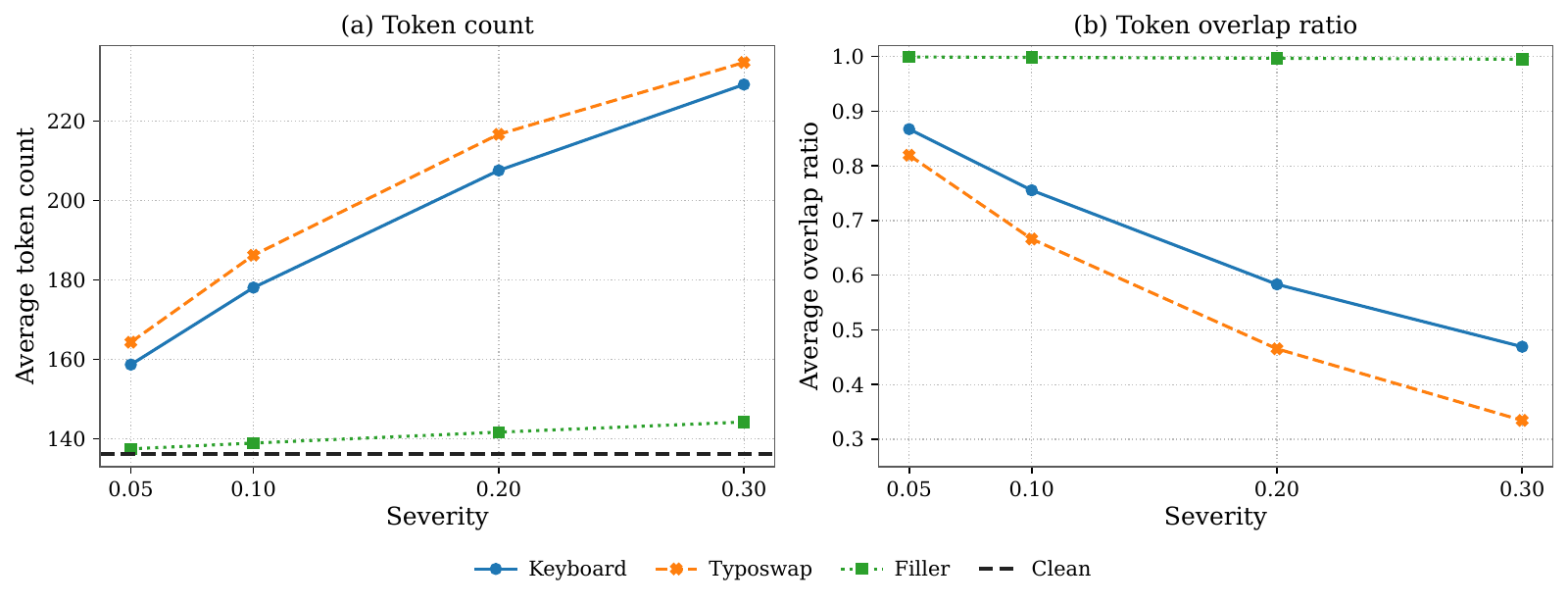}
\caption{\textbf{Tokenization disruption under lexical perturbations.}
(a) Character-level perturbations increase token count with severity, while
filler remains near the clean baseline. (b) Keyboard noise and typoswap
reduce token overlap with clean prompts, indicating token identity
disruption.}
\label{fig:tokenization_disruption}
\end{figure}

Figure~\ref{fig:tokenization_disruption} reports the two tokenization
effects summarized in Section~\ref{sec:mechanism}: character-level
perturbations increase token count with severity while filler remains near
the clean baseline, and keyboard noise and typoswap reduce token overlap
with the clean prompt.

\begin{figure}[!ht]
\centering
\includegraphics[width=\linewidth]{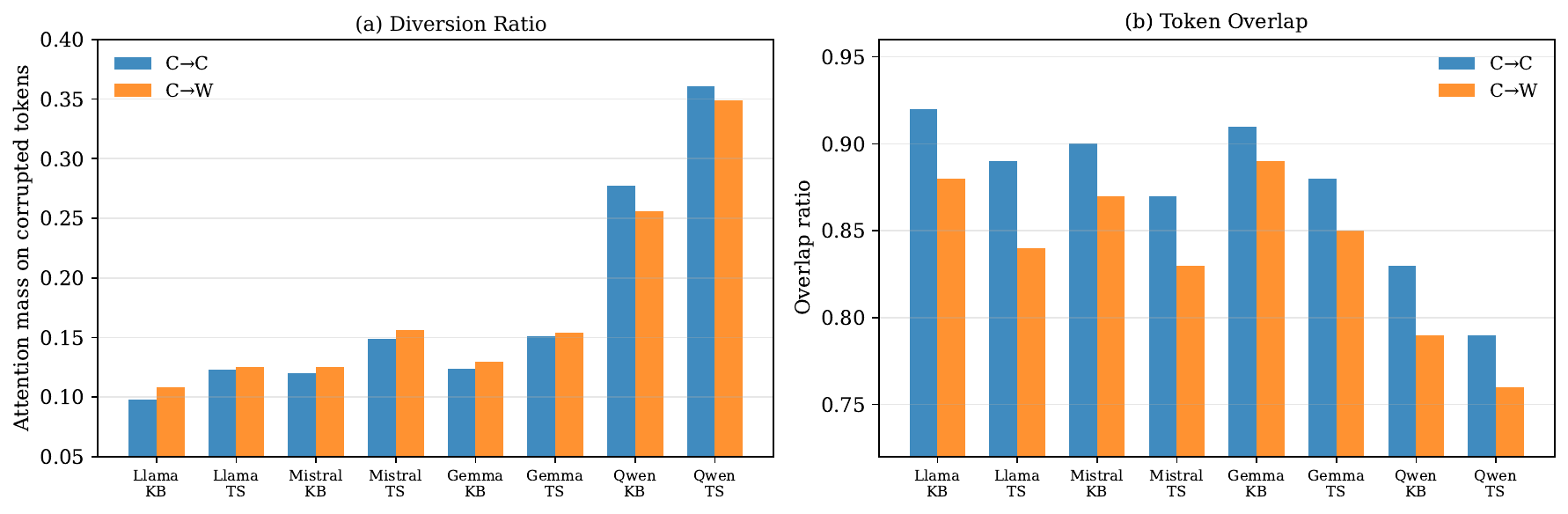}
\caption{\textbf{Error-conditioned evidence for lexical fragility.} Failed
cases (C$\rightarrow$W) exhibit higher diversion ratios and lower token
overlap than cases that remain correct (C$\rightarrow$C).}
\label{fig:error_conditioned_analysis}
\end{figure}

Figure~\ref{fig:error_conditioned_analysis} conditions the same diagnostics
on the outcome, comparing examples that remain correct under perturbation
(\textit{correct$\rightarrow$correct}) against those that flip
(\textit{correct$\rightarrow$wrong}). Failure cases show higher diversion
ratios, lower token overlap, and larger token-count inflation, most clearly
at low-to-moderate corruption levels, linking tokenization disruption and
Attention Diversion to prediction failures rather than to aggregate
statistics alone.

% ====================================================================
\section{Cross-Tokenizer Diversion Analysis}
\label{app:cross_tokenizer}

\begin{figure}[!ht]
\centering
\includegraphics[width=\linewidth]{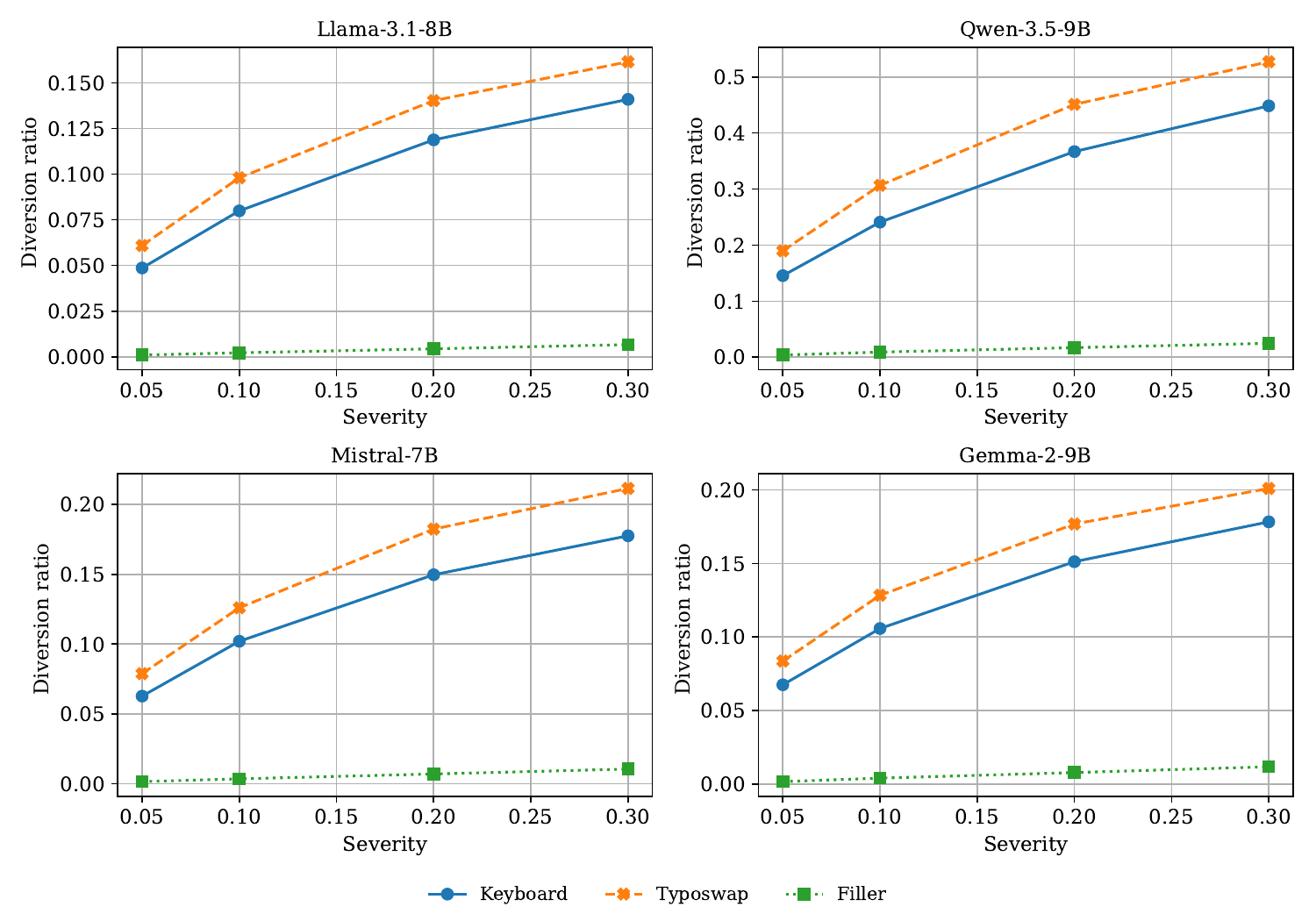}
\caption{\textbf{Attention redistribution toward corrupted tokens.}
Diversion ratio under increasing corruption severity across four models,
computed over all attention layers. Character-level perturbations
consistently increase corrupted-token attention, with typoswap producing
stronger diversion than keyboard noise. Filler insertion stays below 0.025
throughout. Table~\ref{tab:cross_tokenizer} gives the same quantities
numerically.}
\label{fig:attention_hijack_ratio}
\end{figure}

To examine whether tokenizer design affects diversion magnitude,
we report per-model diversion ratios at three severities for keyboard noise and typoswap on
Llama-3.1-8B, Mistral-7B-v0.3, Gemma-2-9B-IT, and Qwen3.5-9B
(Figure~\ref{fig:attention_hijack_ratio} and
Table~\ref{tab:cross_tokenizer}).

\begin{table*}[t]
\centering
\small
\caption{\textbf{Cross-tokenizer diversion ratio at three severities.}
The ranking Qwen $>$ Gemma $>$ Mistral $>$ Llama is consistent across both
perturbation types and all severities, supporting an
architecture/tokenizer-level effect rather than sampling artifact. Shading
marks magnitude only (\colorbox{mag!30}{$\ge$0.20},
\colorbox{mag!14}{0.10--0.20}) and is deliberately distinct from the
degradation shading used elsewhere: Section~\ref{sec:task_resilience} shows
that diversion magnitude does not track accuracy loss.}
\label{tab:cross_tokenizer}
\setlength{\tabcolsep}{6pt}
\begin{tabular}{llcccccc}
\toprule
& & \multicolumn{3}{c}{\textbf{Keyboard ($r$=)}}
  & \multicolumn{3}{c}{\textbf{Typoswap ($r$=)}} \\
\cmidrule(lr){3-5}\cmidrule(lr){6-8}
\textbf{Model} & \textbf{Tokenizer (Vocab)}
  & 0.05 & 0.10 & 0.20 & 0.05 & 0.10 & 0.20 \\
\midrule
Llama-3.1-8B          & tiktoken BPE (128K)   & 0.046 & 0.077 & \mB{0.118} & 0.061 & \mB{0.101} & \mB{0.148} \\
Mistral-7B-v0.3       & SentencePiece (33K)   & 0.058 & 0.096 & \mB{0.146} & 0.075 & \mB{0.122} & \mB{0.181} \\
Gemma-2-9B-IT         & SentencePiece (256K)  & 0.090 & \mB{0.146} & \mA{0.215} & \mB{0.118} & \mB{0.186} & \mA{0.263} \\
Qwen3.5-9B$^{\dagger}$ & tiktoken BPE (152K) & \mB{0.146} & \mA{0.241} & \mA{0.367} & \mB{0.190} & \mA{0.307} & \mA{0.451} \\
\bottomrule
\end{tabular}
\\[2pt]
\footnotesize $^{\dagger}$Qwen3.5-9B uses a hybrid Gated-DeltaNet/Attention
architecture; the comparison conflates tokenizer with architecture, though
the same ordering holds among the three pure transformers.
\end{table*}

Three observations emerge. First, \textbf{vocabulary size is not the
determinant}. Gemma's 256K vocabulary is roughly 8$\times$ Mistral's 33K,
yet Gemma's diversion is consistently higher. Second, \textbf{the ranking
Qwen $>$ Gemma $>$ Mistral $>$ Llama holds across all six reported
conditions} (3 severities $\times$ 2 perturbation types) under the
all-layers diversion metric used throughout the paper, supporting an
architecture or tokenizer-level effect rather than a sampling artifact.
Third, we hypothesize that tokenizer training coverage and merge-rule
differences (multilingual vs.\ English-dominant corpora) may affect how
aggressively English text is split when characters are perturbed. Qwen's
high diversion ratio is also consistent with its large GSM8K typoswap drop
in the main results (Table~\ref{tab:robustness_r01}).

\section{Per-Example Regression Analysis}
\label{app:regression}

Section~\ref{sec:mechanism} reports a logistic regression
quantifying how each input-level signal predicts the
\emph{correct-to-wrong} flip.  This appendix provides the full
regression results, dataset construction, and effect sizes.

\paragraph{Data construction.}
We aggregated per-example records from our four open-weight models
across all four benchmarks, four character-level perturbation rates
($r \in \{0.05, 0.10, 0.20, 0.30\}$), and two perturbation types
(keyboard, typoswap).  We restricted the analysis to examples that
the model solves correctly under clean input ($n=4{,}160$ total),
since these are the cases where the perturbation alone determines
the flip.  The binary outcome is $\textsf{failure} = 1$ if the
clean-correct example becomes wrong under perturbation.

\paragraph{Predictors.} Each example carries four standardized predictors:
fragmentation ratio $\textsf{frag}$, attention diversion ratio $\textsf{diversion}$
(averaged over all attention layers, matching the metric used in the main text),
absolute token-count change $\textsf{n\_delta}$, and token overlap
$\textsf{overlap}$.

\paragraph{Univariate effects.}
Table~\ref{tab:reg_univariate} reports the univariate logistic
regression coefficients with 95\% confidence intervals.  Fragmentation
is the strongest predictor by an order of magnitude.

\begin{table}[t]
\centering
\small
\caption{\textbf{Univariate logistic regression.} Each predictor's standalone
effect on $P(\textsf{failure})$. $\beta$ is the log-odds coefficient per 1 SD
of the predictor, and OR is the corresponding odds ratio. All effects are
highly significant.}
\label{tab:reg_univariate}
\begin{tabular}{lccc}
\toprule
\textbf{Predictor} & $\beta$ & \textbf{OR} & $p$ \\
\midrule
fragmentation & $+0.66$ & 1.93 & $<10^{-81}$ \\
overlap       & $-0.54$ & 0.58 & $<10^{-58}$ \\
n\_delta      & $+0.25$ & 1.29 & $<10^{-14}$ \\
diversion     & $+0.23$ & 1.26 & $<10^{-12}$ \\
\bottomrule
\end{tabular}
\end{table}

\paragraph{Joint model controlling for length.}

To test whether fragmentation's effect is a proxy for prompt-length
inflation, we regress on fragmentation and token-count change jointly
(Table~\ref{tab:reg_joint}). Fragmentation's coefficient
\emph{strengthens} from $\beta{=}0.66$ to $\beta{=}1.04$ after controlling
for length (OR 1.93 to 2.84), while token-count change loses its independent
predictive effect.

\begin{table}[t]
\centering
\small
\caption{\textbf{Joint logistic regression}: fragmentation and token-count
change as simultaneous predictors of failure.  Fragmentation strengthens
when length is controlled, demonstrating that the two signals are not
interchangeable.}
\label{tab:reg_joint}
\begin{tabular}{lccc}
\toprule
\textbf{Predictor} & $\beta$ & \textbf{OR} & $p$ \\
\midrule
fragmentation & $+1.04$ & 2.84 & $<10^{-88}$ \\
n\_delta      & $-0.50$ & 0.61 & $<10^{-23}$ \\
\bottomrule
\end{tabular}
\end{table}

\paragraph{Summary.}
These results provide statistical confirmation of the controlled
fragmentation experiment in Section~\ref{sec:mechanism}: among
input-level signals, fragmentation is the strongest and most robust
predictor of reasoning failure.

\section{Embedding-Replacement Intervention Protocol}
\label{app:embedding_intervention}
Section~\ref{sec:mechanism} reports an embedding-replacement intervention
that directly tests the causal role of fragmented-token embeddings. This
appendix gives the full protocol and per-example diagnostics.

\paragraph{Setup.}
We use Llama-3.1-8B-Instruct loaded in bfloat16 on a single GPU. We evaluate
on 100 GSM8K examples randomly sampled from the test split with seed 42,
under keyboard perturbation at $r=0.1$.

\paragraph{Procedure.}
For each example we run three forward passes. The \textbf{clean baseline}
encodes the clean prompt, decodes greedily, and records accuracy. The
\textbf{corrupted baseline} does the same on the corrupted prompt. The
\textbf{embedding-replaced intervention} encodes the corrupted prompt to
obtain its token-embedding sequence
$\mathbf{e}_{1{:}L_c}^{\textsf{corr}}$ and the clean prompt to obtain
$\mathbf{e}_{1{:}L_d}^{\textsf{clean}}$. For each corrupted token position
$i$ whose token ID does \emph{not} appear in the clean sequence within a
$\pm 5$-position window, we replace its embedding with the most-similar
clean-position embedding (cosine similarity $> 0.3$ within the window). The
resulting hybrid sequence is passed via \texttt{inputs\_embeds} to greedy
decoding.

\paragraph{Results.}
Table~\ref{tab:embedding_intervention_full} reports accuracy and the average
number of token embeddings replaced per example. On average 2.95 embeddings
are replaced out of a corrupted-prompt length of $\approx 75$, and accuracy
recovers from 0.42 to 0.52, restoring 26\% of the 0.38-point gap to clean
accuracy (0.80).

\begin{table}[t]
\centering
\small
\caption{\textbf{Embedding-replacement intervention on GSM8K}
(Llama-3.1-8B, keyboard $r=0.1$, $n=100$).}
\label{tab:embedding_intervention_full}
\begin{tabular}{lc}
\toprule
\textbf{Condition} & \textbf{Accuracy} \\
\midrule
Clean baseline               & 0.80 \\
Corrupted baseline           & 0.42 \\
Embedding replacement        & \textbf{0.52} \\
\midrule
Absolute recovery            & $+$0.10 \\
Fraction of gap recovered    & 26\% \\
Mean embeddings replaced     & 2.95 \\
\bottomrule
\end{tabular}
\end{table}

\paragraph{Interpretation.}
The intervention is conservative: it modifies only embeddings whose
token IDs differ from the clean sequence within a small local window,
and matches each replacement by embedding cosine similarity.  Recovering
26\% of the accuracy gap with $\sim$3 embedding edits per example
provides direct causal evidence that fragmented-token embeddings carry
the load-bearing representational signal that disrupts reasoning,
independent of any downstream attention dynamics. 
This local cosine-matching procedure is the one used for the round-1
intervention reported in Section~\ref{sec:mechanism}. The factorial
experiment in Section~\ref{sec:factorial} uses the difflib alignment described
in Appendix~\ref{app:factorial_protocol} for both of its arms.

%######################################################################

\section{Attention Suppression Intervention}
\label{app:suppression}

\begin{figure}[!ht]
\centering
\includegraphics[width=\linewidth]{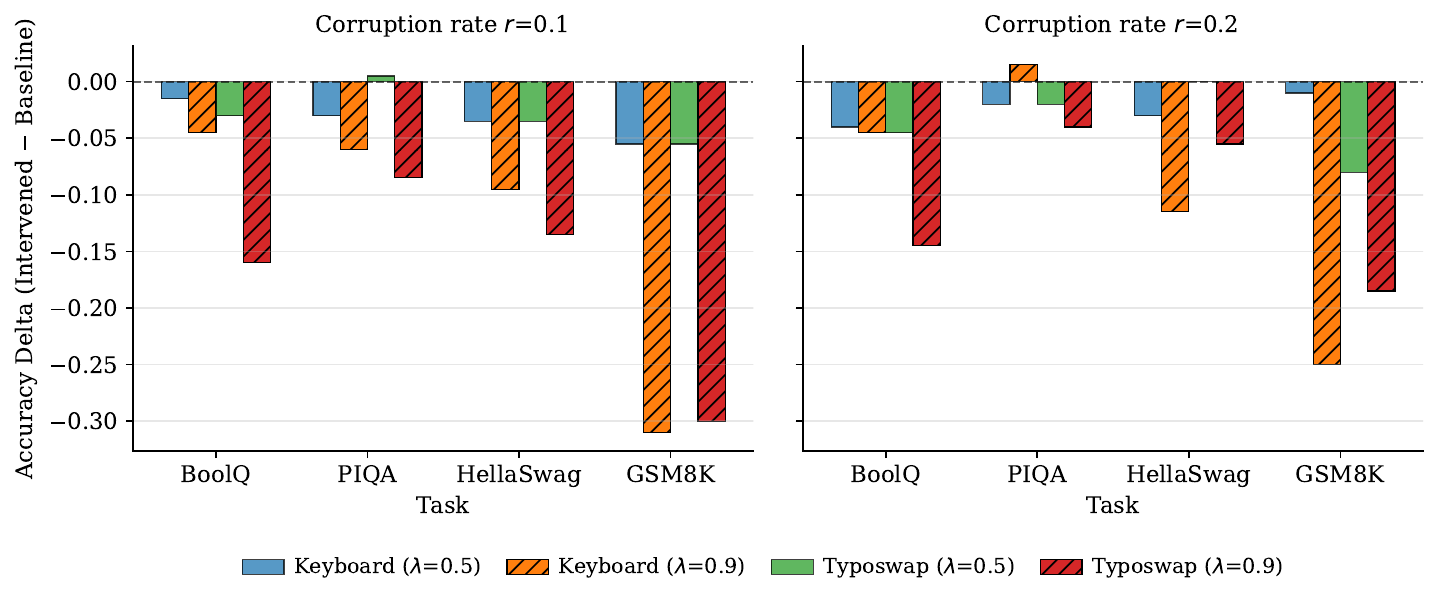}
\caption{\textbf{Intervention via attention suppression.} Suppressing
corrupted-token attention degrades accuracy, indicating that it is
load-bearing rather than passive noise (Llama-3.1-8B, $n{=}200$).}
\label{fig:causal_intervention}
\end{figure}

Section~\ref{sec:mechanism} reports that suppressing attention to outlier
tokens worsens rather than improves accuracy. For each token $j$ we compute
an outlier score $s_j=\|h_j-\bar{h}\|_2$ from the hidden state and rescale
attention as $\tilde{A}_{ij}=A_{ij}\cdot\sigma(-\lambda s_j)$, sweeping
$\lambda\in\{0.5,0.9\}$ on Llama-3.1-8B ($n{=}200$).
Figure~\ref{fig:causal_intervention} shows that stronger suppression
consistently degrades accuracy across tasks, with GSM8K dropping up to
$-$0.31. Corrupted-token attention is therefore load-bearing rather than
passive noise. The factorial intervention of
Section~\ref{sec:factorial} explains why: removing attention from corrupted
positions leaves the model with neither the corrupted evidence nor any
substitute for it.

\section{Factorial Intervention Protocol}
\label{app:factorial_protocol}
Section~\ref{sec:factorial} reports a factorial intervention that crosses the
source of the attention pattern with the source of the token embeddings. This
appendix gives the implementation and the validation checks.

\paragraph{Setup.}
Four open-weight models are evaluated on GSM8K and BoolQ under typoswap at
$r{=}0.1$, with $n{=}100$ examples per model--task cell drawn from a fixed
seed-42 subset, bfloat16 weights, eager attention, and greedy decoding.
Confidence intervals are paired cluster bootstrap over example ids
($B{=}2000$, percentile), with the interaction term computed on the recovery
scale within each replicate. Main-text results report GSM8K. BoolQ's
clean--corrupted gap of 4--10 points makes the recovery ratio ill-conditioned,
so we do not decompose it.

\paragraph{Position alignment.}
Clean and corrupted token-ID sequences are aligned with
\texttt{difflib.SequenceMatcher}, which yields a map from each corrupted
position to a clean counterpart. Within a replaced block, a clean token aligned
to several corrupted fragments maps to each of them. Where a corrupted fragment
has no clean counterpart, the map falls back to the adjacent clean position.
The same map is used for both arms, so the attention and embedding
interventions operate on a common correspondence. Across the evaluation set,
99.3\% of positions receive a genuine clean counterpart and 0.7\% are clamped.

\paragraph{Attention restoration.}
For the clean-attention arm we capture the clean run's pre-softmax
$QK^{\top}/\sqrt{d}$ logits at every attention-bearing layer and head, and write
$S^{\text{restored}}[p,q] = S^{\text{clean}}[\pi(p),\pi(q)] - \log K_{\pi(q)}$,
where $\pi$ is the position map and $K_{\pi(q)}$ is the number of corrupted
fragments aligned to clean position $\pi(q)$. The $-\log K$ term is required for
mass preservation. Without it, a clean token aligned to $K$ fragments has its
logit duplicated across $K$ columns and its post-softmax mass inflated
$K$-fold. With the correction, attention mass on corrupted fragments falls from
0.121 to 0.089 against a clean-run level of 0.079, and without it the same
quantity rises to 0.145. The corrupted run's causal mask and positional
encodings are retained, and the softmax is recomputed after injection.

\paragraph{Embedding restoration.}
For the clean-content arm, the embedding at each corrupted position is replaced
by the embedding of its aligned clean token, using the same map. Because the
hybrid sequence retains the corrupted sequence length, the restoration is
approximate. A clean token aligned to several fragments is written into each of
them, and clean tokens with no corrupted counterpart are dropped. The
restoration is therefore conservative, and if anything understates the content
channel.

\paragraph{Validation.}
Five checks were run before the main grid. \emph{Identity}: with an empty
corruption set the patched forward pass is bit-identical to the unpatched model
(maximum logit difference $0.0$). \emph{Normalization}: post-softmax attention
rows sum to one, matched against the unpatched baseline rather than an absolute
tolerance, since bfloat16 imposes a floor of roughly $3\times10^{-3}$ on
row-sum error. \emph{Causality}: the upper triangle of the attention matrix is
exactly zero after softmax, confirming that the causal mask is applied after
injection rather than before. This matters because the position map can send a
later corrupted position to an earlier clean one. \emph{Coverage}: the fraction
of positions with a genuine clean counterpart is reported above.
\emph{Mass}: attention mass on corrupted fragments moves toward the clean-run
level, as reported under attention restoration.

% ====================================================================
\section{Fragmentation Rates Across Perturbations and Languages}
\label{app:fragmentation_rates}

This appendix provides the per-perturbation fragmentation rates
that support the cross-perturbation analysis in
Section~\ref{sec:mechanism}. We report mean fragmentation (the ratio
of new tokens introduced by perturbation to clean-prompt token count)
and token overlap (the fraction of clean-prompt token types preserved).
Because the cross-perturbation controls require matched ASR/OCR variants, the
English diagnostic subset used for Table~\ref{tab:frag_70b_english} is
separate from the 70B generalization subset reported in
Appendix~\ref{app:scale_lang}. Small differences in clean accuracy therefore
reflect sampling variation across $n=100$ subsets.

\paragraph{English (BoolQ + GSM8K, Llama-3.3-70B tokenizer).}
Table~\ref{tab:frag_70b_english} in the main text reports fragmentation
under six perturbation types at $r=0.1$--$0.2$, alongside the
corresponding accuracy from Section~\ref{sec:mechanism}.

\paragraph{Filler vs.\ Keyboard across four English tokenizers.}
To verify that filler's negligible impact on accuracy is grounded
in low fragmentation, we measured fragmentation rates across four
tokenizers (Llama-3.1, Mistral-7B-v0.3, Gemma-2-9B, Qwen3.5-9B)
on BoolQ+GSM8K averaged.  Table~\ref{tab:frag_filler_vs_keyboard}
reports mean fragmentation at four severities.  Keyboard perturbation
produces 2--3$\times$ higher fragmentation than filler at matched $r$,
quantitatively explaining the asymmetric accuracy patterns in
Section~\ref{sec:robustness_results}.

\begin{table}[t]
\centering
\small
\caption{\textbf{Mean fragmentation across four open-weight tokenizers},
BoolQ+GSM8K averaged ($n=100$ each task).
Keyboard fragmentation exceeds filler by 2--3$\times$ at matched
word-level perturbation rates.}
\label{tab:frag_filler_vs_keyboard}
\begin{tabular}{lcc}
\toprule
\textbf{$r$} & \textbf{Filler} & \textbf{Keyboard} \\
\midrule
0.05 & 0.006 & 0.013 \\
0.10 & 0.015 & 0.039 \\
0.20 & 0.029 & 0.097 \\
0.30 & 0.048 & 0.147 \\
\bottomrule
\end{tabular}
\end{table}

\paragraph{Chinese (CMath, Llama-3.1 tokenizer).}
Table~\ref{tab:frag_cmath} reports fragmentation under Chinese
keyboard, typoswap, and filler perturbations at three severities.
The ordering matches English: keyboard $>$ typoswap $>$ filler,
with filler producing fragmentation an order of magnitude below
keyboard at all severities.

\begin{table}[t]
\centering
\small
\caption{\textbf{Chinese CMath fragmentation rates}
(Llama-3.1 tokenizer, $n=200$).}
\label{tab:frag_cmath}
\begin{tabular}{lccc}
\toprule
\textbf{Perturbation} & $r{=}0.05$ & $r{=}0.10$ & $r{=}0.20$ \\
\midrule
keyboard & 0.049 & 0.100 & 0.198 \\
typoswap & 0.024 & 0.047 & 0.080 \\
filler   & 0.010 & 0.026 & 0.050 \\
\bottomrule
\end{tabular}
\end{table}

%##########################################################################

\section{Qualitative Case Study}
\label{app:case_study}

\begin{figure*}[!t]
\centering
\includegraphics[width=\textwidth]{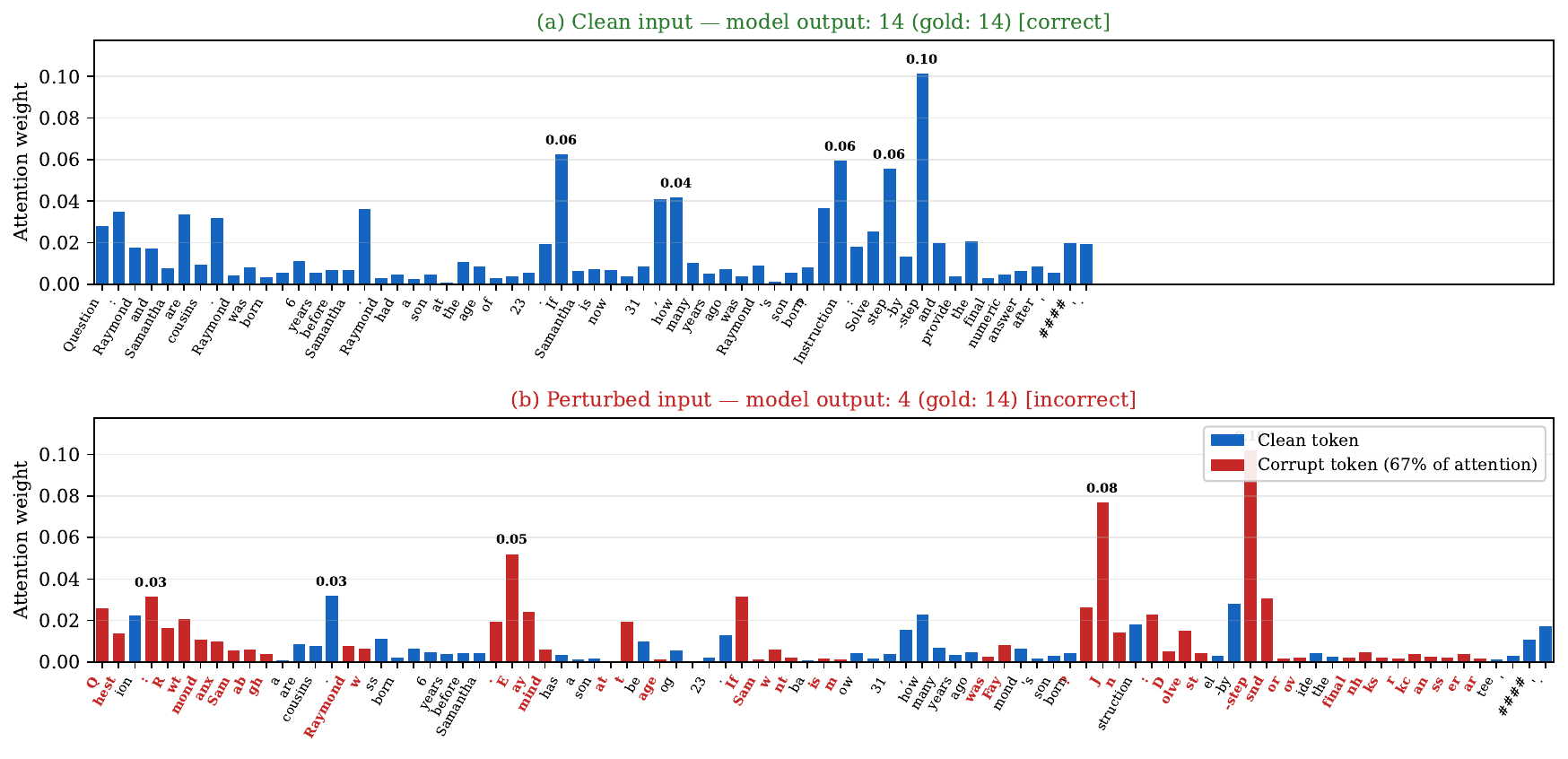}
\caption{\textbf{Qualitative attention analysis on a GSM8K failure case}
(Llama-3.1-8B, keyboard perturbation, $r{=}0.1$, layers 12--15).
\textbf{(a)}~Under clean input, attention concentrates on
semantically relevant tokens such as names, ages, and instruction
keywords, yielding the correct answer~14.
\textbf{(b)}~After perturbation, 67\% of the attention mass
shifts to corrupted subword fragments (red bars), while
task-critical tokens are starved of processing bandwidth.
The model outputs~4 instead of~14.  Token count increases
from 48 to 63 due to tokenization fragmentation.}
\label{fig:case_study}
\end{figure*}

Figure~\ref{fig:case_study} presents a per-token attention comparison
for a GSM8K example that Llama-3.1-8B answers correctly under clean input
but incorrectly after keyboard perturbation at $r{=}0.1$.
After perturbation, tokenization fragmentation increases the token count
from 48 to 63, and corrupted fragments absorb 67\% of attention mass,
leaving only 33\% for task-relevant tokens.

\section{Attention Calibration and Prompt Restoration}
\label{app:diagnostic_probes}
Section~\ref{sec:mitigation} notes that attention calibration and prompt
restoration show no consistent improvement. This appendix reports those probes.

\paragraph{Setup.}
Both probes were run on Llama-3.1-8B at $r{=}0.1$ with $n{=}100$, on a subset
drawn by \texttt{shuffle(seed=42)} and using an independently implemented
corruption function. The subset and prompt template therefore differ from the
repair experiments in Section~\ref{sec:mitigation}, so the baselines are not
comparable across the two. Each is internally paired.
Table~\ref{tab:diagnostic_probes} reports both probes.

\paragraph{Calibration.}
Attention calibration rescales attention mass away from tokens flagged as
outliers by representation norm, leaving the input unchanged. It yields
negligible or negative changes in all three conditions tested.

\paragraph{Restoration.}
Prompt restoration rewrites flagged tokens toward their nearest in-vocabulary
form before inference. It degrades accuracy in all three conditions, most
sharply on BoolQ, where it miscorrects valid passage tokens.

\begin{table}[t]
\centering
\small
\caption{\textbf{Diagnostic probes on Llama-3.1-8B} ($n{=}100$, $r{=}0.1$;
shuffle-42 subset with its own corruption implementation, so baselines differ
from Section~\ref{sec:mitigation}).}
\label{tab:diagnostic_probes}
\setlength{\tabcolsep}{5pt}
\begin{tabular}{@{}lccc@{}}
\toprule
\textbf{Condition} & \textbf{Baseline} & \textbf{Calibration} & \textbf{Restoration} \\
\midrule
BoolQ + KB     & 64.0 & 61.0 & 53.0 \\
BoolQ + TS     & 83.0 & 82.0 & 65.0 \\
HellaSwag + KB & 22.0 & 18.0 & 19.0 \\
\bottomrule
\end{tabular}
\end{table}

\paragraph{Interpretation.}
Neither probe recovers accuracy, and both are consistent with the coupling result
of Section~\ref{sec:factorial}. Calibration adjusts attention while leaving the
corrupted content in place, which the factorial intervention shows to be
ineffective at best. Restoration modifies content without knowing which tokens
were corrupted, and its errors propagate in the same way as the confident-but-wrong
substitutions analysed in Section~\ref{sec:defense}.

% ====================================================================
\section{Tokenizer-Level Defense Protocol}
\label{app:defense_protocol}

Section~\ref{sec:defense} reports a detection-and-repair defense operating without
access to the clean text. This appendix gives the protocol.

\paragraph{Constraint.}
The detector and the repair observe only the corrupted input. The perturbation
mask is used solely to score detector precision and recall after the fact, in a
code path never called during detection or repair. All thresholds and the deployed
detector were selected on a held-out split disjoint from the evaluation subset.

\paragraph{Detection.}
Three signals are combined. Rarity flags words that are out of vocabulary or fall
below a unigram-frequency threshold. Surprisal flags tokens whose local negative
log-likelihood under a small reference model exceeds a threshold. Fragmentation
flags words whose subword-pieces-per-character ratio is unusually high under the
target tokenizer. We evaluate each signal alone, their union, and a majority vote,
and additionally a high-precision variant requiring both the majority vote and
out-of-vocabulary status.

\paragraph{Repair.}
Vocabulary-constrained correction replaces each flagged word with the highest-frequency
in-vocabulary word within edit distance two. A conservative variant restricts this
to edit distance one and declines to act when no candidate with non-zero frequency
exists. Character-level fallback, which re-encodes flagged spans as per-character
tokens, and greedy longest-match retokenization were implemented and dry-run but
not deployed: both rewrite every flagged word and therefore inherit the detector's
false-repair rate of 22--25\%.

\paragraph{Evaluation.}
Four open-weight models on GSM8K and BoolQ, typoswap at $r{=}0.1$, $n{=}100$ per
model--task cell. Recovery is reported as a percentage of the clean--corrupted gap
with paired bootstrap confidence intervals. 
Every variant is additionally run on
clean inputs, so that a defense which improves corrupted accuracy at the cost of
clean accuracy can be identified. No variant shows significant clean-input
degradation.
% Every variant is additionally run on
% clean inputs, so that a defense which improves corrupted accuracy at the cost of
% clean accuracy can be identified; no variant shows significant clean-input
% degradation.

\paragraph{Results.}
Detection reaches F1 0.77--0.81 with the majority-vote detector, and fragmentation
alone reaches 0.70--0.75 without any clean text. Vocabulary-constrained correction
restores the original word in 67.0\% of truly corrupted words at edit distance two
and 69.9\% under the conservative variant. Recovery of the accuracy gap is
significantly positive in none of the eight model--task cells and significantly
negative in one.

\section{Extended Related Work}
\label{app:related_work}

Section~\ref{sec:related_work} organizes prior work by where each line
locates the remedy for lexical corruption: at the input, in the attention
pattern, or during training. This appendix gives the full citations and
expands each comparison with the evidence behind it.

\paragraph{Robustness to textual perturbations.}
Character-level noise and spelling errors have long been known
to degrade neural NLP systems,
particularly in machine translation and sequence labeling
\cite{belinkov2018synthetic,pruthi2019combating}.
Adversarial attack methods such as TextBugger \cite{li2019textbugger}
and HotFlip \cite{ebrahimi2018hotflip} demonstrate that
small surface-form modifications can substantially alter predictions
even when semantic meaning is preserved.
Related work on reading comprehension shows that
neural models often rely on brittle lexical cues
\cite{jia2017adversarial}.
These studies primarily target earlier architectures
(LSTMs, BERT-scale transformers) and search for the edit that maximally
changes an output, whereas our perturbations are undirected and sampled
uniformly over words, so the degradation we report is the expected cost of
ordinary typing and transcription noise rather than a worst case.
They also locate the remedy at the input, correcting the surface form before
the model reads it. Our cross-perturbation controls delimit when that
succeeds. 
ASR homophones produce no fragmentation and leave accuracy at
baseline, and OCR confusions fragment more heavily than any other
perturbation yet still leave accuracy near baseline, because their pieces
follow familiar subword patterns. Only typoswap, whose fragments are out of
distribution, damages GSM8K (Table~\ref{tab:frag_70b_english}).
Correction
therefore works exactly where the original form is still recoverable from
vocabulary and context, and fails on the corruptions that dominate
mathematical reasoning, where a perturbed numeral remains a valid,
in-vocabulary numeral admitting no dictionary neighbour. The tokenizer-level
defense makes the bound concrete: detection reaches F1 0.77--0.81 without
access to clean text, yet vocabulary-constrained repair restores the original
word only 67--70\% of the time and recovers no accuracy in any of eight
model--task cells (Section~\ref{sec:defense}).
A separate line intervenes during training instead. Subword regularization,
in particular BPE-Dropout \cite{provilkov2020bpe}, randomizes segmentation so
that a model never comes to depend on a single canonical tokenization. Our
mechanism supplies the reason this is the effective point of intervention:
once fragmentation has occurred, the information needed to undo it is no
longer present in the input, so any later stage is working from a
representation that has already lost what it would need.

\paragraph{LLM reasoning and evaluation.}
Reasoning benchmarks such as BoolQ \cite{boolq}, PIQA \cite{piqa},
HellaSwag \cite{hellaswag}, and GSM8K \cite{gsm8k}
are widely used to evaluate LLM capabilities.
Chain-of-thought prompting \cite{wei2022chain}
has been shown to improve multi-step reasoning,
and subsequent work explores self-consistency \cite{wang2023selfconsistency}
and other inference-time strategies.
Recent studies also examine LLM sensitivity to prompt formatting
\cite{sclar2024quantifying,lu2022fantastically},
showing that minor template variations can affect performance.
In all of these settings the prompt varies while the tokenization of its
content stays intact, so the failure is one of framing and better prompting
can address it. Lexical corruption changes the token identities themselves,
which is why stronger reasoning does not help: chain-of-thought lifts clean
GSM8K on GPT-4o from 0.42 to 0.90, confirming the reasoning capability is
intact, yet typoswap still pulls it to 0.70, and GPT-5.4's built-in reasoning
mode shows the largest keyboard degradation of any frontier configuration we
evaluate, amplifying the damage rather than absorbing it
(Appendix~\ref{app:frontier_models}).

\paragraph{Attention analysis and mechanistic interpretability.}
Attention patterns are widely used to analyze
transformer behavior \cite{vaswani2017attention,clark2019does}.
Recent mechanistic interpretability work examines
how specific attention heads contribute to reasoning
\cite{geva2021transformer,elhage2021mathematical}.
Studies on attention sinks \cite{xiao2024efficient}
show certain tokens attract disproportionate attention
regardless of relevance.
Attention Diversion describes a related but distinct phenomenon:
the absorbing positions are not fixed properties of the model but are created
by the input itself, as corrupted fragments become transient sinks that draw
attention mass away from task-relevant tokens.
The difference matters for what can be done about it. Static sinks can be
suppressed or offset, whereas input-induced sinks cannot be ablated without
cost, since the fragments that absorb attention are also what the model uses
to reconstruct the intended word. Two results establish this. Rescaling
attention away from outlier tokens degrades accuracy in every condition
tested, by up to $0.31$ on GSM8K (Appendix~\ref{app:suppression}), so the
diverted attention is load-bearing rather than passive noise. The factorial
intervention then shows it is not independently manipulable either: supplying
a clean attention pattern over corrupted content is significantly harmful in
three of four models, restoring content alone reaches significance in only
one, and only restoring both together recovers 45--71\% of the gap, with a
significantly positive interaction in every model
(Section~\ref{sec:factorial}). Diversion is therefore neither a passive
symptom to be corrected nor a free-standing cause to be suppressed, but is
functionally bound to the corrupted content it addresses.

\end{document}